\pdfoutput=1

\documentclass[11pt]{article}

\usepackage[final]{acl}

\usepackage{times}
\usepackage{latexsym}

\usepackage[T1]{fontenc}

\usepackage[utf8]{inputenc}
\usepackage{amssymb}

\usepackage{microtype}

\usepackage{inconsolata}

\usepackage{graphicx}

\usepackage{subfig}
\usepackage{xcolor}

\usepackage{multirow}  
\usepackage{threeparttable}  
\usepackage{colortbl}  
\usepackage{rotfloat}  
\usepackage{array}  
\usepackage{booktabs}  
\usepackage{amsmath}
\usepackage{amsfonts}
\usepackage{amssymb}
\usepackage{adjustbox}
\usepackage{algorithm}
\usepackage{algorithmic}
\usepackage{array}

\usepackage{listings}

\definecolor{Gray}{gray}{.95}
\definecolor{Blue}{RGB}{240, 250, 255}
\definecolor{blue2}{RGB}{83, 125, 201}
\definecolor{pink2}{RGB}{236, 101, 98}
\definecolor{purple}{RGB}{103, 99, 181}
\title{TailorKV: A Hybrid Framework for Long-Context Inference via Tailored KV Cache Optimization}

\author{
  Dingyu Yao$^{1,2*}$ , Bowen Shen$^{1,2}$\ , Zheng Lin$^{1,2}$\footnotemark[2] , Wei Liu$^{3}$, Jian Luan$^{3}$, Bin Wang$^{3}$,\\
  \bf Weiping Wang$^{1}$\\
  $^{1}$Institute of Information Engineering, Chinese Academy of Sciences, Beijing, China\\
  $^{2}$School of Cyber Security, University of Chinese Academy of Sciences, Beijing, China\\
  $^{3}$MiLM Plus, Xiaomi Inc, Beijing, China\\
  \texttt{\{yaodingyu, shenbowen, linzheng, wangweiping\}@iie.ac.cn}\\
  \texttt{\{liuwei40, luanjian, wangbin11\}@xiaomi.com}
}

\begin{document}
\maketitle
\begin{abstract}
  \renewcommand{\thefootnote}{\fnsymbol{footnote}}
  \footnotetext[1]{\ \ Work done during an internship at Xiaomi Inc.}
  \footnotetext[2]{\ \ Corresponding Author: Zheng Lin.}
  \renewcommand{\thefootnote}{\arabic{footnote}}

The Key-Value (KV) cache in generative large language models (LLMs) introduces substantial memory overhead. Existing works mitigate this burden by offloading or compressing the KV cache. However, loading the entire cache incurs significant latency due to PCIe bandwidth bottlenecks in CPU-GPU communication, while aggressive compression causes notable performance degradation. We identify that certain layers in the LLM need to maintain global information and are unsuitable for selective loading. In contrast, other layers primarily focus on a few tokens with dominant activations that potentially incur substantial quantization error. This observation leads to a key insight that loading dominant tokens and quantizing all tokens can complement each other. Building on this insight, we propose a hybrid compression method, TailorKV, which seamlessly integrates quantization and offloading. TailorKV develops an inference framework along with a hardware-friendly implementation that leverages these complementary characteristics. Extensive long-context evaluations exhibit that TailorKV achieves nearly lossless performance under aggressive compression settings, outperforming the state-of-the-art. Particularly, the Llama-3.1-8B with 128k context can be served within a single RTX 3090 GPU, reaching 82 ms per token during decoding\footnote{Code is available at: \url{https://github.com/ydyhello/TailorKV}.}.

\end{abstract}

\section{Introduction}

Large language models (LLMs) have demonstrated exceptional performance in tasks such as multi-turn dialogues~\cite{chiang2023vicuna} and multi-document understanding~\cite{bai-etal-2024-longbench}. In response to the growing complexity of tasks, recent LLMs have expanded their context windows to over 128k tokens, e.g., GPT-4~\cite{achiam2023gpt} and DeepSeek V3~\cite{liu2024deepseek}.
Typically, the inference of LLMs is auto-regressive, with the Key-Value (KV) cache stored in memory to avoid recomputation. However, the size of KV cache grows linearly with sequence length, leading to much higher GPU memory consumption and inference latency.

Recent studies have proposed sparse attention mechanisms to reduce KV cache usage. These methods fall into two categories: irreversible eviction and recallable selection. 
Irreversible eviction methods~\cite{li2024snapkv,zhang2023h2o,xiao2024efficient} suffer from \textbf{accuracy degradation} due to permanently discarding tokens that may later become crucial, particularly in multi-turn dialogues.
Recallable selection methods adopt a different approach by maintaining the entire KV cache while selecting only a subset of tokens for processing. However, methods like Quest~\cite{tang2024quest} and SparQ~\cite{ribar2024sparq} encounter \textbf{memory limitations} when attempting to store all tokens on the GPU.
Although CPU offloading mitigates GPU memory limitations, existing approaches~\cite{xiao2024infllm,zhang2024pqcache} still require retrieving a substantial portion of tokens (around 20\%), introducing significant \textbf{decoding latency overheads} due to slow data transfer between CPU RAM and GPU RAM.

To optimize accuracy, memory, and latency simultaneously, we first analyze the compression preferences for the KV cache based on layer characteristics.
Prior researches~\cite{feng2024ada,cai2024pyramidkv} applied different sparsity rates to different layers under the same compression strategy.
However, our analyses demonstrate that performance degradation primarily stems from the application of unsuitable compression at the layer level (Section~\ref{insight:spareserror}). 
Therefore, we suggest that shallow layers, which exhibit dense attention patterns and emphasize global information~\cite{wan2025textdtexto}, are better suited for uniform compression like quantization.
Conversely, layers with a few dominant tokens and largely redundant information are well-suited for sparsity, as performance can be maintained by retrieving only the dominant tokens.

Building upon these insights, we propose a novel framework, \textbf{TailorKV}, which employs hybrid compression techniques to reduce GPU memory usage. We introduce an identification metric to classify Transformer layers into two distinct types: \textbf{\textit{(i)}} 
\textit{quantization-friendly} layers, which preserve global information from a macro perspective, and \textbf{\textit{(ii)}} \textit{sparsity-friendly} layers, which capture crucial information from a micro perspective.
This design enables 
\textit{quantization-friendly} layers to employ static quantization, achieving a high compression ratio (1-bit per floating-point number) while maintaining model quality. Meanwhile, for \textit{sparsity-friendly} layers, the system offloads the KV cache to CPU memory during prefilling and dynamically retrieves the Top-K tokens during decoding. 
By aligning compression strategies with the characteristics of each layer, this tailored approach significantly reduces overall memory consumption.

The accuracy and efficiency of TailorKV are evaluated on various backbone LLMs using long-context benchmarks. The results demonstrate that TailorKV drastically reduces memory usage by quantizing 1 to 2 layers to 1-bit precision and loading only 1\% to 3\% of the tokens for the remaining layers while maintaining nearly lossless performance.
Specifically, TailorKV achieves a decoding latency of 82 ms for Llama-3.1-8B with a 128k-context on a single RTX 3090 (PCIe 1.0)\footnote{We combine TailorKV with 4-bit weight-only quantization~\cite{lin2024awq} for prefill phase memory allocation.}, yielding a 53.7\% reduction in peak GPU memory usage. The key contributions are summarized as follows.

\begin{itemize}
\item We identify layer-specific compression preferences and develop an identification metric to determine optimal compression strategies for different layers in the model.
\item We present TailorKV, a hybrid KV cache compression framework that combines quantization and offloading techniques through an algorithm-system co-design, preserving both model accuracy and execution efficiency.
\item Extensive experiments on long-context benchmarks demonstrate the nearly lossless performance of TailorKV with minimal GPU memory consumption and acceptable latency.
\end{itemize}

\section{Preliminaries}

\begin{figure*}[t]
\centering
\subfloat[Dense (left) and sparse (right) attention.]{
    \label{fig: attention_score}
    \includegraphics[width=0.76\columnwidth]{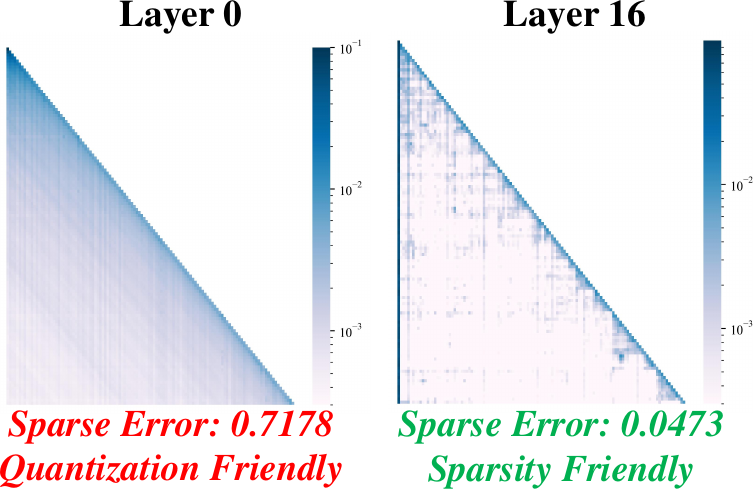}
}
\hspace{0.05em}
\subfloat[Sparse error on different LLMs.]{
    \label{fig: Sparse_Error_Model} \includegraphics[width=0.59\columnwidth]{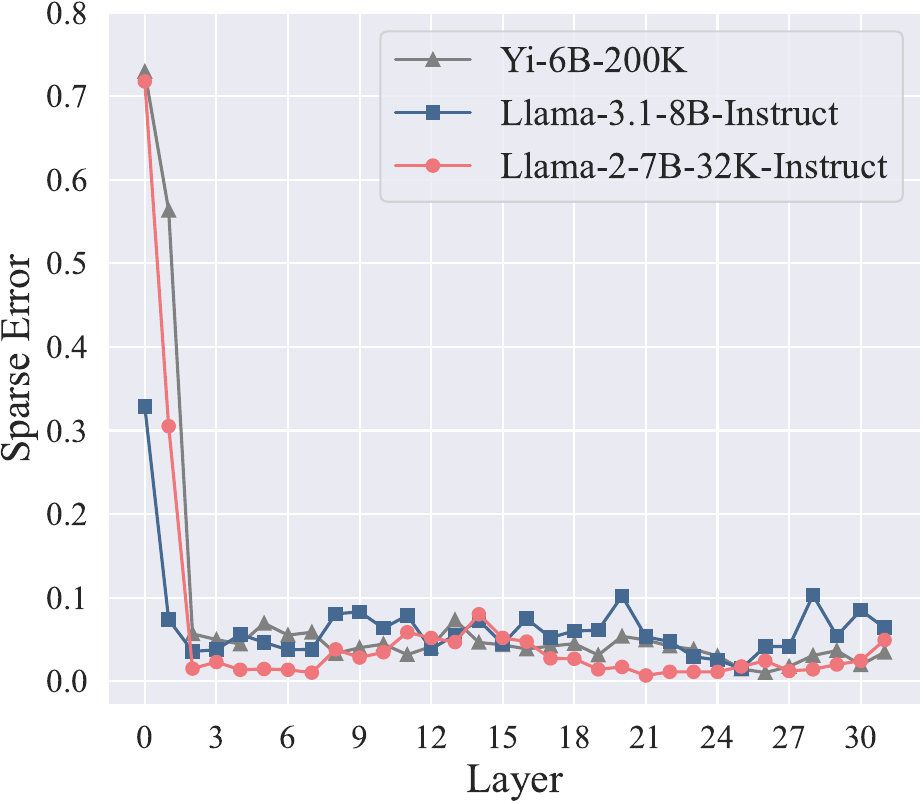}
}
\hspace{0.05em}
\subfloat[Sparse error on different datasets.]{
    \label{fig: Sparse_Error_LongBench}
    \includegraphics[width=0.61\columnwidth]{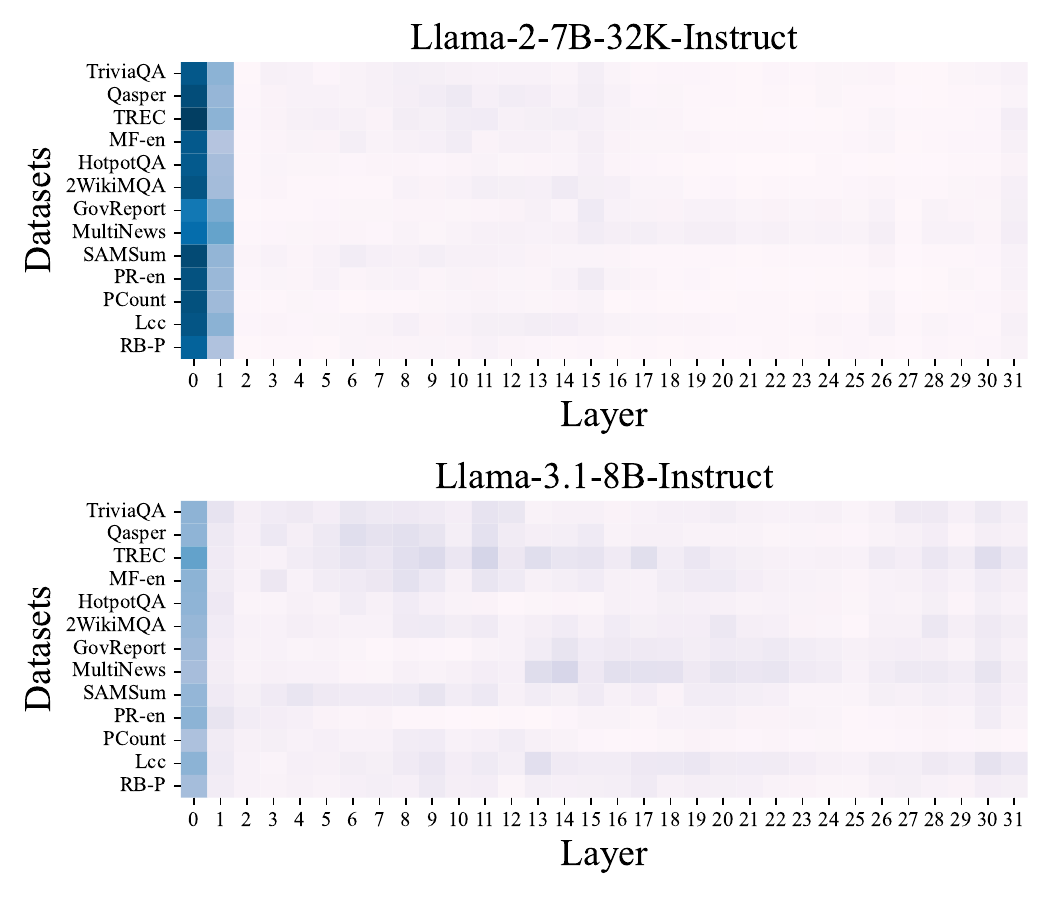}
}
  \caption{\textbf{Observations on attention.} (a) Attention weights on Llama-2-7B-32K-Instruct. Detailed visualizations are in \autoref{appendix:detail_attention}. (b) Sparse error of different models on the 2WikiMQA dataset, with only the top 5\% of attention scores retained. (c) Sparse error on different datasets, with only the top 5\% of attention scores retained.}
  \label{fig:motivation}
\end{figure*}

\subsection{Attention and KV Cache}

LLM inference consists of two stages: \textbf{\textit{prefill}} and \textbf{\textit{decode}}. During prefilling, the entire prompt is used to generate the first token. Consider the prompt embedding \( \mathbf{X} \in \mathbb{R}^{n \times d} \) along with the weight matrices \( \mathbf{W}_i^q, \mathbf{W}_i^k, \mathbf{W}_i^v \in \mathbb{R}^{d \times  d_h} \) for head \(i \in [1,h]\), where \( n \) is the sequence length, \(d\) is the hidden dimension and \( d_h \) is the head dimension. The keys and values for head \(i\) are computed and cached, as follows:
\begin{equation}
\mathbf{K}_i= \mathbf{X W}_i^k,~\mathbf{V}_i= \mathbf{X W}_i^v.
\end{equation}

During decoding, the new token embedding \( \mathbf{x} \in \mathbb{R}^{1 \times d}\) is computed iteratively to produce the query, key, and value vectors. The cache is updated and the output \( \mathbf{o} \) of each attention head is computed as:
\begin{equation}
    \begin{aligned}
      \mathbf{K}_i \gets  \mathrm{Cat}[\mathbf{K}_i,\mathbf{x} \mathbf{W}_i^k],
     \mathbf{V}_i \gets \mathrm{Cat}[\mathbf{V}_i,\mathbf{x} \mathbf{W}_i^v],
    \end{aligned}
\end{equation}
\begin{equation}
    \begin{aligned}
    \mathbf{a}_i = \mathrm{Softmax}\bigl( \mathbf{q}_i \mathbf{K}_i^\top / \sqrt{d_h}\bigr), \mathbf{o}_i =\mathbf{a}_i\mathbf{V}_i,
    \end{aligned}
    \label{eq:selfattention}
\end{equation}
where \(\mathbf{q}_i=\mathbf{x}\mathbf{W}_i^q\), and the attention outputs from all heads are concatenated and sent to the FFN.

\subsection{Quantization of KV Cache}
Quantization converts continuous or high-precision values into lower-precision discrete representations. 
Given a tensor \(\mathbf{X}\) in high precision, the typical uniform quantization process can be expressed as:
\begin{equation}
\begin{aligned}
\mathbf{X}_Q=&~\mathrm{Quant}_b(\mathbf{X},s,z)\\
=&~\mathrm{clamp}(\lfloor \frac{\mathbf{X} - z}{s} \rceil,0,2^b-1),
\label{eq:quant}
\end{aligned}
\end{equation}
where \(\mathbf{X}_Q\) represents the quantized tensor in \(b\)-bit precision, with \(z=\min \mathbf{X}\) as the zero-point and \( s=\frac{\max \mathbf{X}-\min \mathbf{X} }{2^b - 1} \) as the scaler. The clamp function restricts values to the \(b\)-bit integer range and $\lfloor\cdot\rceil$ denotes the rounding function.

\subsection{GPU-CPU Co-execution}
As the sequence length increases, the size of the KV cache grows, significantly raising the demand for GPU resources. For example, with a sequence length of 512k, Llama-2-7B~\cite{touvron2023llama} requires up to 256GB of memory for the KV cache. Current LLM serving systems~\cite{kwon2023efficient,qin2025mooncake} employ an offloading strategy that stores the KV cache in cost-effective CPU memory and loads it onto the GPU during inference. However, I/O transfer latency becomes the bottleneck in inference due to the low-bandwidth PCIe interface~\cite{zhang2024dovetail}.
For instance, transferring the KV cache of a single layer (\(\approx\)~8GB) from the CPU memory to the RTX 3090 GPU via PCIe 1.0 link (4GB/s) takes around 2s, while the attention computation for a single layer on the RTX 3090 GPU only takes around 10ms.
Thus, on-demand fetching is currently the most common approach to reduce GPU idle time.

\section{Motivations and Observations}

\label{insight:spareserror}
\begin{table}[t]
\centering
\small
\setlength{\tabcolsep}{3pt}

\begin{tabular}{lcccc}
\toprule
\textbf{Strategy} & \textbf{RB-P} & \textbf{LCC} & \textbf{GovReport} & \textbf{TriviaQA}  \\
\midrule
 16-bit & 56.7 & 63.4 & 34.9 & 91.6 \\
1-bit (\textit{KIVI}) & 24.4 & 26.2 & 8.3 &18.6 \\
\midrule
1-bit (\(\mathbb{L} = \{0\}\)) & \textbf{57.1} & \textbf{62.6} & \textbf{34.9} & \textbf{92.1} \\
1-bit (\(\mathbb{L} =  \{2\}\)) & 53.0 & 59.3 & 32.6 & 91.9 \\
1-bit (\(\mathbb{L} = \{10\}\)) & 52.3 & 59.3 & 31.2 & 90.7 \\
1-bit (\(\mathbb{L} = \{18\}\)) & 53.7 & 60.0 & 34.9 & 89.4 \\
\bottomrule
\end{tabular}
  \caption{Results of 1-bit quantization on different layers, using Llama-3.1-8B. \(\mathbb{L}\) denotes the quantized layer. \textit{KIVI}~\cite{liu2024kivi} is an advanced KV cache quantization algorithm. } 
  \label{tab:1bit}
\end{table}

\paragraph{Layers have compression preferences.}
In contrast to previous belief~\cite{li2024snapkv,zhang2023h2o}, we propose that not all layers are suitable for sparsity.  
To quantify the sparsity challenges during decoding, we define the sparse error \(\mathcal{E}\). Let \(\mathbf{a} \in \mathbb{R}^{1 \times  n}\) represents the attention weight as defined in \autoref{eq:selfattention}, and let  \(\mathcal{M}\in \{0,1\}^n\) denote a binary mask that selects the top \(k\) elements of \({\mathbf{a}}\). The sparse error \(\mathcal{E}\) for each head is defined as:
\begin{equation}
\begin{aligned}
\hat{\mathbf{a}} = \mathbf{a} \odot \mathcal{M},~~\mathcal{E}= 1 -\begin{matrix} \sum_{i=1}^{n} \end{matrix}\hat{\mathbf{a}}_{i}.
\end{aligned}
\end{equation}
As shown in Figure\autoref{fig: attention_score}, layers with dense attention distributions exhibit higher sparse errors compared to those with sparse distributions.
Additionally, we observe sparse error patterns across models and datasets. 
Figure\autoref{fig: Sparse_Error_Model} shows that sparse errors are similar across models, with higher sparse errors in shallower layers (e.g., 0, 1). Figure\autoref{fig: Sparse_Error_LongBench} shows that the distribution of sparse errors remains consistent across various datasets for the same model. 

\begin{figure}[t]
  \centering
  \includegraphics[width=1\linewidth]{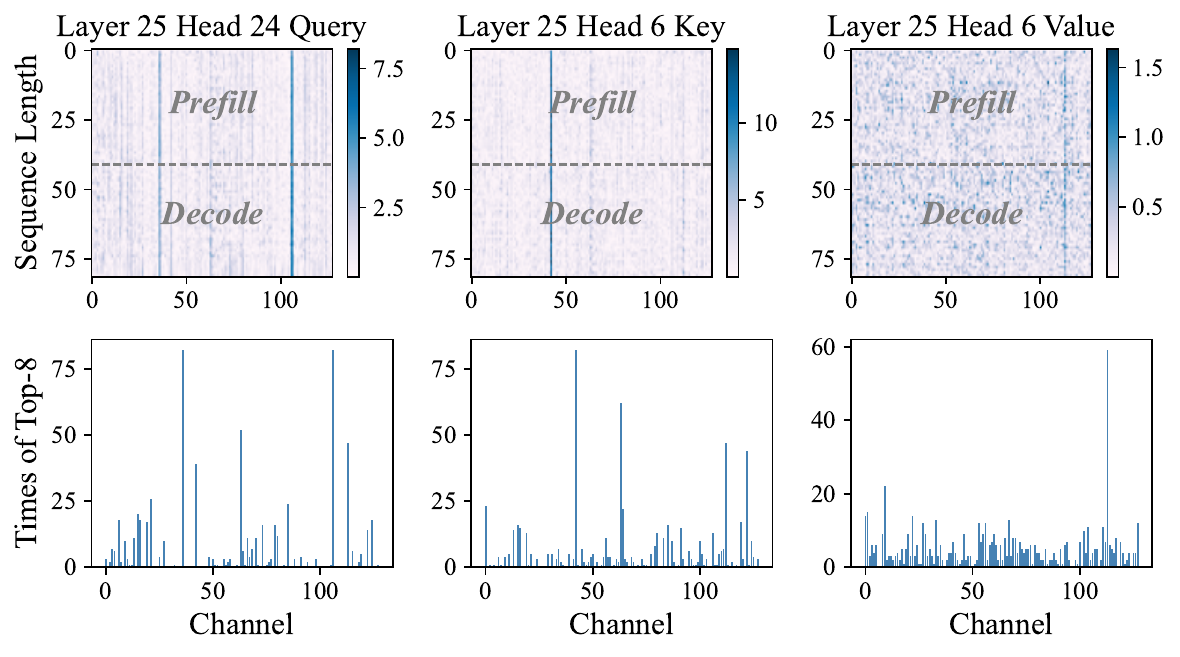} 
  \caption{(Top) Query and key in Llama-3.1-8B-Instruct show outlier patterns in some channels, while the value shows no outliers. (Bottom) The number of times reaching the Top-8. Outliers may appear in any position.}
  \label{fig:motivation2}
\end{figure}
Similarly, not all layers are suitable for quantization.
As shown in \autoref{tab:1bit}, quantizing the KV cache to 1-bit leads to significant performance degradation. This degradation is primarily caused by layers with sparse distributions, which are more sensitive to quantization. In contrast, quantizing the dense layer (e.g., 0th) incurs no performance loss.

These findings highlight the need for a tailored KV cache compression strategy.
We regard layers with dense distributions as \textbf{\textit{quantization-friendly}}, which focus on global information, and layers with sparse distributions as \textbf{\textit{sparsity-friendly}}, which prioritize crucial information. 
\begin{figure*}[t]
  \centering
  \includegraphics[width=1\linewidth]{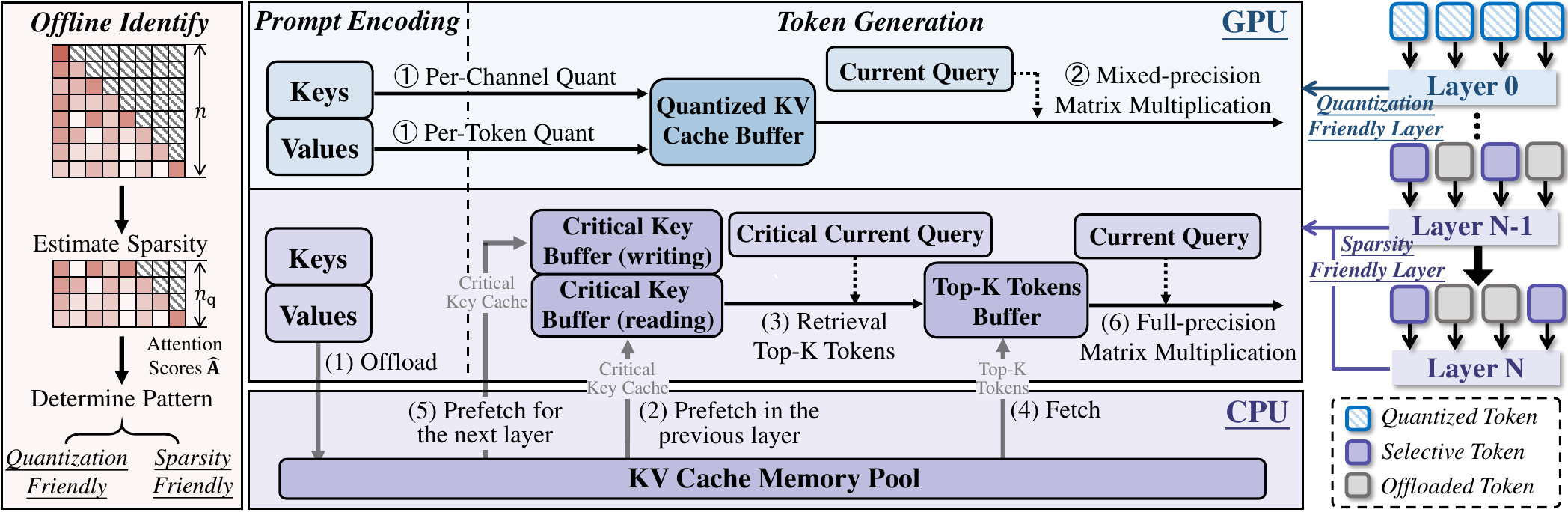} 
  \caption{System overview of \textbf{TailorKV}. \textcolor{pink2}{Offline identification} categorizes the layers into quantization-friendly and sparsity-friendly. For quantization-friendly layers, we employ aggressive \textcolor{blue2}{static quantization}. For sparsity-friendly layers, we \textcolor{purple}{dynamically retrieve} Top-K tokens. Critical current query and critical key cache represent the outliers in the query and key cache, respectively.}
  \label{fig:overview}
\end{figure*}
\paragraph{Attention scores correlate with outliers.}
\label{insight:channel_outliers}

Each channel in the key and query contributes to the attention scores through their dot product, as expressed by the formula \( \mathbf{qK^\top} \). \autoref{fig:motivation2} (Top) illustrates that some channels have large magnitudes in the query and key. It follows from the dot product formula that attention scores correlate with these outliers.
A recent method~\cite{yang2024post} focuses on static channel sparsity, utilizing offline calibration technique to identify high-magnitude channels.
However, we find that the sparsity of query and key channels is dynamic rather than static. As shown in \autoref{fig:motivation2} (Bottom), outliers in the query and key do not consistently appear in fixed positions; instead, they may appear in any position. Furthermore, dynamically selecting high-magnitude channels improves the recall of dominant tokens compared to using a static offline strategy. This claim is empirically validated in Section~\ref{sec:ablation}.


\section{Methodology}

\subsection{Offline Identification}

Empirical observation in Section~\ref{insight:spareserror} suggests that some layers benefit more from quantization, while others are better suited for sparsity.
To avoid disrupting the standard inference, we apply an offline strategy to identify the compression preference of each layer. 
In this phase, we introduce a metric—dense preference score \( \mathcal{P} \)—to assess whether each attention layer favors quantization or sparsity. 
Given a prompt length \(n\), we first use the most recent \(n_q\) query vectors \( \mathbf{Q_{\text{last\_q}}} \in \mathbb{R}^{n_q \times  d_h} \) and the key vectors \( \mathbf{K} \in \mathbb{R}^{n \times d_h } \) to compute the attention score matrix \( \hat{\mathbf{A}}\) for each head during prefilling:
\begin{equation}
\hat{\mathbf{A}} = \mathrm{Softmax}\bigl(\mathbf{Q_{\text{last\_q}}}\mathbf{K}^{\top} / \sqrt{d_h}\bigr). 
\end{equation}

Next, we select the top \(k\) indices from \(\hat{\mathbf{A}}\) and sum the top \(k\) elements in order to compute the dense preference score \( \mathcal{P} \):
\begin{equation}
\begin{aligned}
\hat{\mathcal{I}} = \left\{ (i,j)\mid\mathrm{Top}_k(\hat{\mathbf{A}}_{i,:}, k) \right\}_{i=1}^{n_q},
\end{aligned}
\end{equation}
\begin{equation}
\begin{aligned}
\mathcal{P} = n_q- \begin{matrix}\sum_{(i,j) \in \hat{\mathcal{I}}} \end{matrix} \hat{\mathbf{A}}_{i,j}.
\end{aligned}
\end{equation}
If the dense preference score \( \mathcal{P}_l \) of layer  \( l \) exceeds the threshold \( \tau \), the layer is regarded as quantization-friendly; otherwise, it is deemed sparsity-friendly. This can be formalized as:
\begin{equation}
\small
C(l)=
\begin{cases} 
\texttt{Quantization-Friendly}, & \text{if } \mathcal{P}_l > \tau , \\
\texttt{Sparsity-Friendly}, & \text{otherwise}.
\end{cases}
\end{equation}

The threshold \( \tau \) is a predefined hyperparameter, and its optimal value is determined through experimentation on the synthetic Longbench task.
The metric \(\mathcal{P}\) consistently assesses the same model across various datasets (for details, see \autoref{appendix:offline}). After layer-level identification, we apply \textbf{\textit{dynamic retrieval}} for sparsity-friendly layers and \textbf{\textit{static quantization}} for quantization-friendly layers. The overall workflow is shown in \autoref{fig:overview}.

\subsection{Dynamic Retrieval}

\begin{figure}[t]
  \includegraphics[width=1\linewidth]{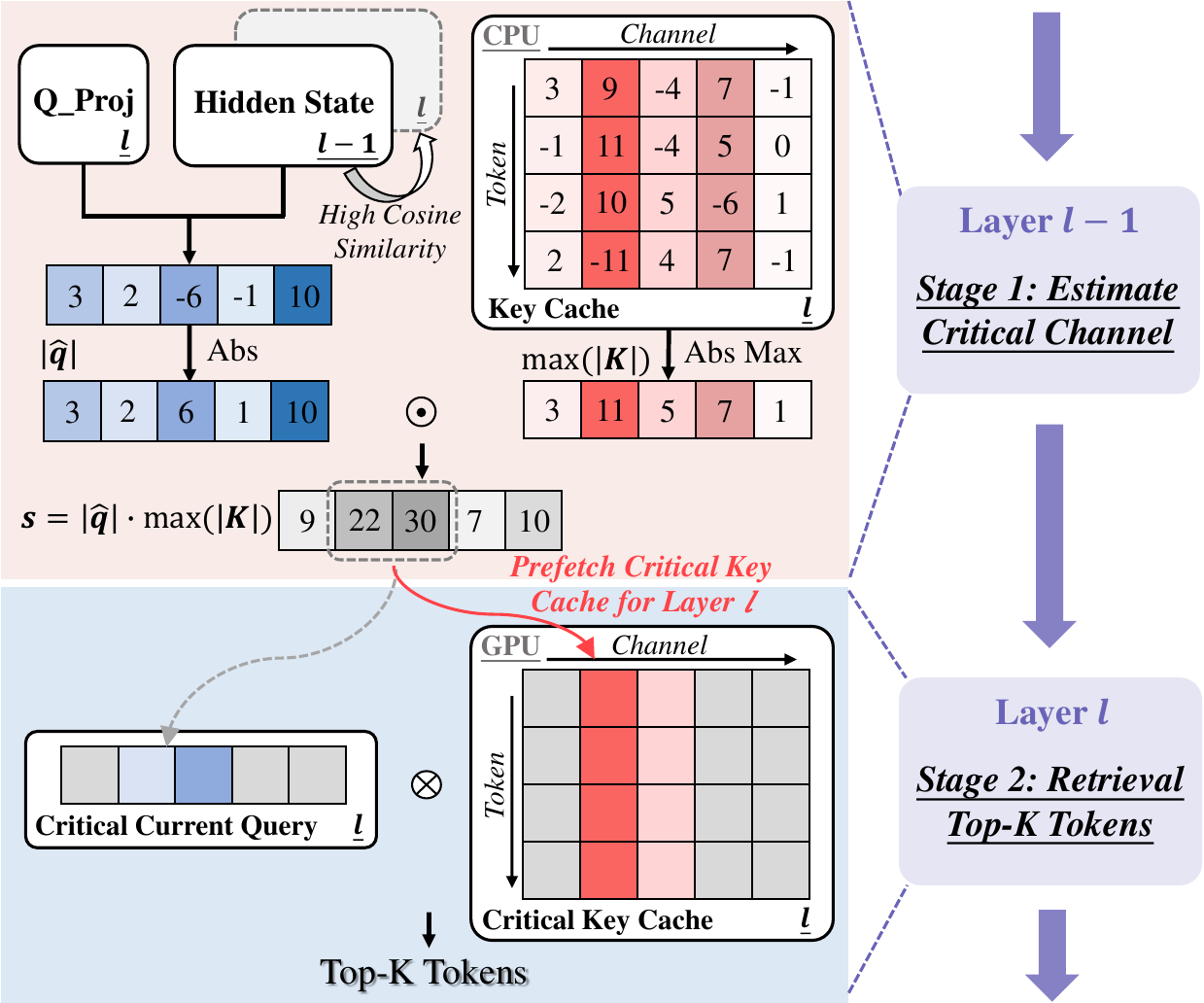} 
  \caption {Two-stage dynamic retrieval process: \textbf{Stage 1} estimates critical channels at layer \(l-1\) and prefetches critical key cache for layer \(l\). \textbf{Stage 2} approximates attention scores and selects Top-K tokens at layer \(l\).}
    \label{fig:select}
\end{figure}

For sparsity-friendly layers, we propose a dynamic retrieval algorithm with an asynchronous system design. 
\autoref{fig:overview} shows the management framework of the CPU memory pool and GPU memory buffer.
To facilitate LLM inference on memory-limited devices, we offload the KV cache to lower-cost CPU memory layer by layer during prefilling. 
Subsequently, we retrieve the Top-K tokens on demand during decoding, thus minimizing communication overhead. 
The core design is illustrated in \autoref{fig:select}.

As explained in Section~\ref{insight:channel_outliers}, attention scores correlate with outliers in the query and key. To more accurately assess token importance, we approximate attention scores prior to original operation based on this insight. We first \textbf{estimate the critical channels} to identify outliers in the query and key cache, referred to as the critical current query and critical key cache. Since the critical key cache resides in the CPU, we employ prefetching to load it in advance. We leverage inter-layer similarity to predict the critical channels ahead of time (for a detailed explanation, see \autoref{appendix:similarity}). The similarity between adjacent layers arises from the residual connection, as validated in prior research~\cite{lee2024infinigen}. 
At layer \( l-1 \), we estimate the query \( \hat{\mathbf{q}} \) for layer \( l \), using the weight matrix from layer \( l \) and the hidden state from layer \( l-1 \). The contribution of the $i$-th channel to the attention scores is computed via element-wise multiplication of $\hat{\mathbf{q}}$ and $\mathbf{K}$:
\begin{equation}
\mathbf{s}_i=\left| \hat{\mathbf{q}}_i \right| \cdot \max (\left| \mathbf{K} _i\right|),i=1,2,...,d_h.
\label{eq:channel}
\end{equation}

\begin{figure}[t]
  \centering
  \includegraphics[width=1\linewidth]{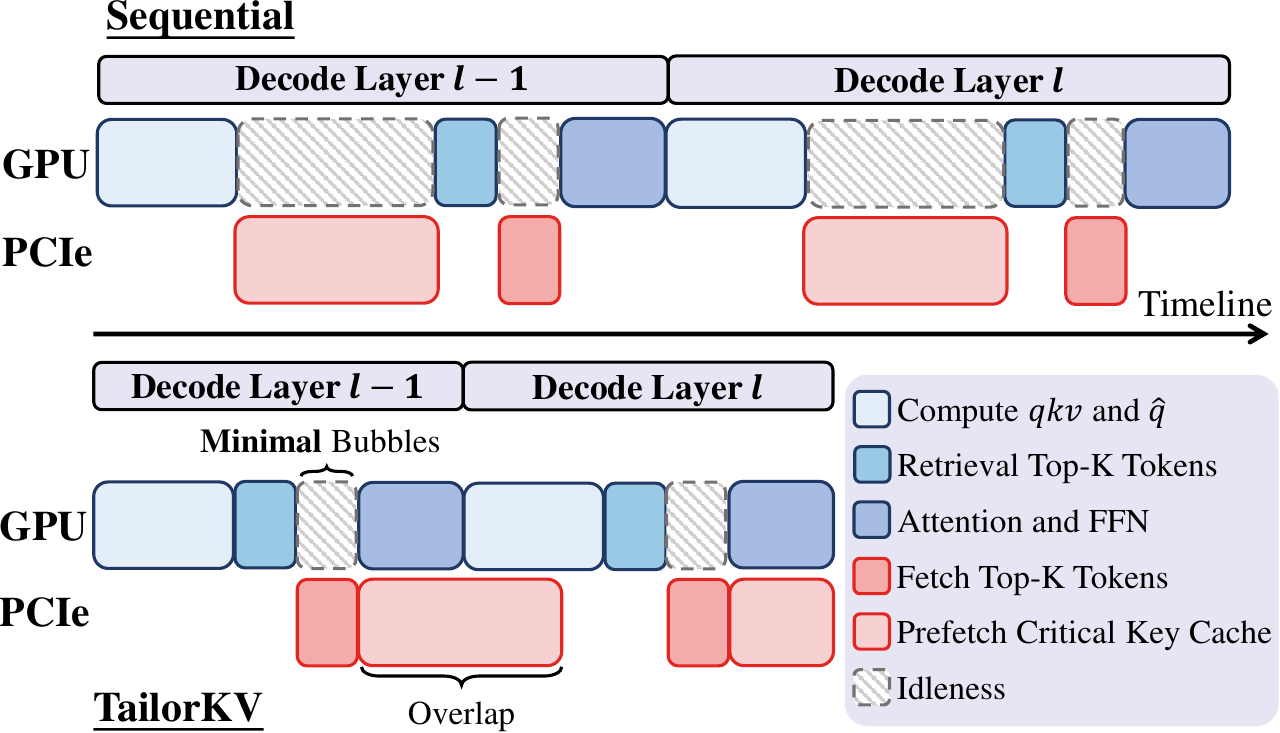} 
  \caption{Timeline of dynamic retrieval. \textcolor{blue2}{Blue} signifies computation and \textcolor{pink2}{pink} signifies communication.
}
  \label{fig:schedule}
\end{figure}

Next, we prefetch the \(l\)-th layer's critical key cache based on \(\mathbf{s}\), using double buffering—one buffer for writing and the other for reading—to enable concurrent execution. Then, we \textbf{retrieve the Top-K tokens} by approximating the attention scores at \(l\)-th layer based on the critical current query and the critical key cache, followed by fetching the Top-K tokens.
\autoref{fig:schedule} outlines the computation and communication during decoding. The only non-overlappable operation is fetching Top-K tokens, as it depends on the current layer's query. TailorKV demonstrates how a heterogeneous design overcomes resource constraints by leveraging CPU-GPU co-execution.

\subsection{Static Quantization}

Unlike traditional quantization methods~\cite{liu2024kivi,yang2024no,he2024zipcache}, TailorKV focuses on ensuring that each layer "plays its role," thus enabling more aggressive compression scheme, such as 1-bit quantization.
As illustrated in \autoref{fig:motivation2}, outliers are present in the key cache along the channel dimension, while the value cache contains no outliers. For quantization-friendly layers, we apply static per-token quantization to the value cache and static per-channel quantization to the key cache~\cite{liu2024kivi}. 
As shown in \autoref{eq:quant}, we introduce a 1-bit quantization kernel and also implement FP16\(\times\)INT1 GEMV to improve hardware performance under aggressive compression.

\subsection{Memory Footprint Analysis}

Let the number of layers be \( L \), the number of heads be \( h\), the sequence length be \( n \), and the head dimensions be \( d_h \). All input tokens are represented in FP16. We present a comparison of GPU memory usage across different methods in \autoref{tab:memory_analysis}.
TailorKV mainly manages a \textbf{quantized KV cache buffer} in quantization-friendly layers and a \textbf{critical key buffer} in sparsity-friendly layers. The quantization zero-point and scaler are stored in FP16 format.

\begin{table}[h]
\centering
\small
\begin{tabular}{lcp{3.01cm}}
\toprule
\textbf{Method} & \textbf{Memory} & \textbf{Parameters} \\
\midrule

 Original & \(2Lnhd_h\) &  -\\

SnapKV & \(2\alpha Lnhd_h \)   & budget: \(\alpha\)\\
Quest & \(2 Lnhd_h (1+ \frac{1}{\beta})\)   & page size: \(\beta\)\\
\midrule
Ours ($\mathbb{Q}$) & \(2l_qnhd_h( \frac{1}{16} + \frac{2}{g} )\)  &  bit size: 1, group size: \( g \),  num layers of $\mathbb{Q}$: \(l_q\) \\
Ours ($\mathbb{S}$) & \(2nhd_s\) & num critical channel: \(d_s\) \\
\bottomrule
\end{tabular}
  \caption{Comparison of memory usage among different methods. The symbols $\mathbb{Q}$ and $\mathbb{S}$ denote the quantization-friendly layer and sparsity-friendly layer, respectively. } 
  \label{tab:memory_analysis}
\end{table}

\begin{table*}[t]
        \centering
        \small
        \setlength{\tabcolsep}{1.7pt}
        \renewcommand{\arraystretch}{1}
        
        \begin{tabular}{@{}l|cccccccc|ccccccc@{}}
            \toprule
            \multirow{3}{*}{\textbf{Methods}} & \multicolumn{8}{c}{\textbf{LongBench}} & \multicolumn{7}{c}{\textbf{InfiniteBench}} \\
            \cmidrule(l){2-9}\cmidrule(l){10-16}
            & \textbf{Tokens} & \textbf{SD.QA} & \textbf{MD.QA} & \textbf{Summ} & \textbf{FS.L} & \textbf{Code} & \textbf{Synth} & \textbf{Avg.} & \textbf{Tokens} & \textbf{Retr} & \textbf{Dia} & \textbf{Novel} & \textbf{Math} & \textbf{Code} & \textbf{Avg.} \\
            \midrule
            \rowcolor{Blue}
            \textit{Llama-3.1-8B} & 128k & 49.6 & 50.9 & 31.2 & 69.4 & 60.0 & 53.5 & 53.8 & 128k & 99.6 & 19.0 & 30.2 & 34.0 & 22.8 & 44.0\\
            StreamLLM & 192 & 26.3 & 42.7 & 17.9 & 50.0 & 48.2 & 53.5 & 40.6 & 1024 & 3.2 & 7.0 & 23.7 & 34.0 & 22.8 & 18.3\\
            SnapKV & 192 & 35.2 & 48.1 & 20.2 & 56.5 & 52.8 & 52.5 & 45.2 & 1024 & \textbf{96.6} & 9.5 & 27.4 & 34.0 & 22.8 & 41.0\\
            Quest & 192 & 40.1 & 46.9 & 20.7 & 61.6 & 48.0 & 52.4 & 46.2 & 1024 & 64.4 & 14.0 & 25.7  & 34.0 & \textbf{25.1} & 33.8 \\
            PQCache & 192 & 48.4 & 49.5 & 27.0 & 67.3 & 56.3 & \textbf{53.6} & 51.7 & 1024 & 5.5 & 15.0 & 27.5 & 34.0 & 23.3 & 21.5\\
            \rowcolor{pink!20}
            \texttt{TailorKV-1} & 64(+128) &  48.2 & \textbf{50.9} & 29.2 & 68.1 & \textbf{58.3} &53.4  &  52.6& 128(+896) & 86.5 & 18.0 & 28.9 & \textbf{34.0} & 22.8 & 40.4\\
            \rowcolor{pink!20}
            \texttt{TailorKV-2} & 64(+128) & \textbf{49.3} & 50.5 & \textbf{29.4} & \textbf{68.7} & 58.1 & 53.3 & \textbf{52.9} & 128(+896) & 94.8 & \textbf{18.5} & \textbf{30.0} & \textbf{34.0} & 22.8 & \textbf{42.6}\\ \midrule
            \rowcolor{Blue}
            \textit{Yi-9B} & 200k & 36.6 & 44.7 & 28.8 & 60.6 & 69.6 & 35.0 & 47.0 & 200k & 100.0 & 2.5 & 25.2 & 23.4 & 26.3 & 39.2\\
            StreamLLM & 192 & 21.3 & 33.6 & 11.0 & 44.1 & 51.8 & 14.7 & 30.6 & 1024 & 1.5 & 2.5 & 24.2 & 23.7 & 21.3 & 16.4\\
            SnapKV & 192 & 25.0 & 38.8 & 11.9 & 49.0 & 59.7 & 18.8 & 35.0 & 1024 & 59.0 & 3.0 & 24.9 & 22.5 & \textbf{26.6} & 30.0\\
            Quest & 192 & 29.2 & 37.9 & 15.4 & 57.5 & 59.6 & 25.7 & 39.1 & 1024 & 98.4 & 4.0 & 21.8 & 18.2 & 18.7 & 36.1\\
            PQCache & 192 & 32.4 & 41.6 & 19.2 & 58.6 & 64.4 & \textbf{27.8} & 42.0 & 1024 & 7.8 & 2.0 & 25.3 & 22.2 & 25.6 & 18.5\\
            \rowcolor{pink!20}
            \texttt{TailorKV-1} & 64(+128) & \textbf{38.0} & \textbf{44.3} & \textbf{27.3} & \textbf{60.2} & \textbf{66.3} & 24.3 & \textbf{44.7} & 128(+896) & \textbf{98.7} & 2.5 & \textbf{26.6} & \textbf{24.0} & 21.3 & 39.2\\
            \rowcolor{pink!20}
            \texttt{TailorKV-2} & 64(+128) & 35.6 & 43.5 & \textbf{27.3} & 60.1 & 66.0  & 23.5 &44.0 & 128(+896) & 98.5 & \textbf{4.5} & 25.3 & \textbf{24.0} & 24.9 & \textbf{39.4}\\
            \midrule
            \rowcolor{Blue}
            \textit{Yi-6B} & 200k & 32.4 & 15.3 & 1.3 & 49.9 & 69.8 & 9.5 & 29.7 & 200k & 99.2 & 0.0 & 25.2 & 6.8 & 26.9 & 37.1\\
            StreamLLM & 192 & 20.0 & 11.6 & \textbf{1.6} & 34.0 & 44.6 & 4.0 & 20.4 & 1024 & 2.0 & 0.0 & 21.8 & 4.8 & 25.8 & 13.5\\
            SnapKV & 192 & 24.2 & 13.0 & \textbf{1.6} & 38.5 & 51.2 & 3.7 & 23.3 & 1024 & 55.7 & 2.5 & 23.2 & 4.5 & \textbf{26.9} & 26.4\\
            Quest & 192 & 26.5 & 12.5 & 0.3 & 46.9 & 51.9 & \textbf{8.5} & 26.2 & 1024 & \textbf{99.5} & 3.0 & 22.0 & 5.1 & 26.6 & 35.8\\
            PQCache & 192 & 30.4 & 14.5 & 0.6 & 48.0 & 55.8 & 4.0 & 27.3 & 1024 & 6.9 & 1.5 & 24.6 & 5.7 & \textbf{26.9} & 16.3\\
            \rowcolor{pink!20}
            \texttt{TailorKV-1} & 64(+128) & \textbf{32.5} & \textbf{15.4} & 1.4 & \textbf{49.7} & 55.9 & 4.0 & \textbf{28.3} & 128(+896) & 98.7 & 2.5 & \textbf{25.3} & 7.7 & 26.4 & 37.2\\
            \rowcolor{pink!20}
            \texttt{TailorKV-2} & 64(+128) & \textbf{32.5} & 15.3 & 1.5 & 49.1 & \textbf{56.4}  & 4.0 &28.2 & 128(+896) & 98.5 & \textbf{3.0} &\textbf{25.3} &\textbf{8.0} & 26.7 & \textbf{37.3}\\
            \bottomrule
        \end{tabular}
        \caption{Task performance (\%) on \textbf{LongBench} and \textbf{InfiniteBench}. 13 sub-tasks of LongBench are aggregated into 6 classes, and 9 sub-tasks of InfiniteBench are aggregate into 5 classes. The aggregation of sub-tasks is discussed in \autoref{tab:longbench_detail} and \autoref{tab:InfiniteBench_detail}, while the detailed results for all sub-tasks can be found in \autoref{tab:longbench} and \autoref{tab:infinitebench}.}
        \label{tab:acc}
\end{table*}
\begin{figure*}[t]
  \centering
  \includegraphics[width=1\linewidth]{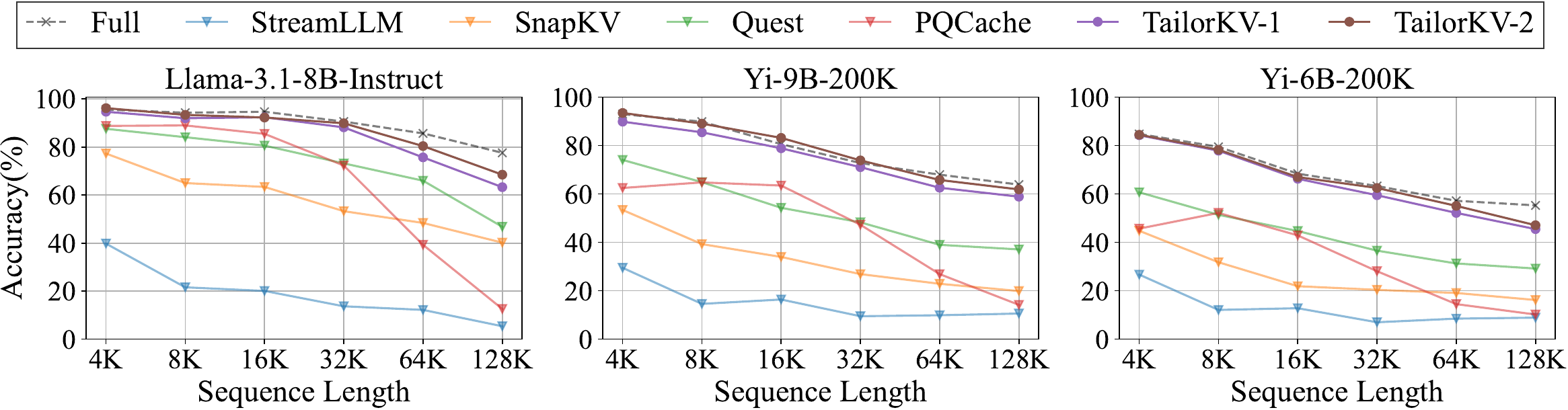} 
  \caption{The average accuracy of different methods on \textbf{RULER}. The sparsity-friendly layer in TailorKV uses 128+(896) tokens, while other methods use 1024 tokens. See \autoref{tab:appendix_ruler} for details. }
  \label{fig:ruler}
\end{figure*}
\section{Experiments}
\subsection{Experimental Setup}
\paragraph{Baselines and Benchmarks.}
We evaluate three widely used models with their respective context lengths: Llama-3.1-8B-Instruct~\cite{dubey2024llama}, Yi-6B-200K~\cite{yi6b}, and Yi-9B-200K~\cite{yi9b}.
To demonstrate the superior performance of our method, we compare TailorKV with competitive baselines, including StreamingLLM~\cite{xiao2024efficient}, SnapKV~\cite{li2024snapkv}, Quest~\cite{tang2024quest}, and PQCache~\cite{zhang2024pqcache}.
To evaluate the performance in long-context scenarios, we employ three well-designed benchmarks, including LongBench~\cite{bai-etal-2024-longbench}, InfiniteBench~\cite{zhang-etal-2024-bench}, and RULER~\cite{hsieh2024ruler}. Refer to \autoref{appendix:evaluating_tasks} for further details.

\paragraph{Implementation.}
We set \(\tau\) to 0.2 for all models.
\texttt{TailorKV-1} and \texttt{TailorKV-2} represent KV cache stored with 1-bit and 2-bit precision in quantization-friendly layers, respectively. The group size is 64, with the zero point and scaler stored in 16-bit.
For sparsity-friendly layers, the number of tokens involved in attention computation is \(n_{\text{local}} + (n_{\text{topk}})\), where \(n_{\text{local}}\) refers to the GPU budget and \(n_{\text{topk}}\) represents the additional communication overhead.
The number of critical channels is 8 for LongBench and 12 for both InfiniteBench and RULER.
The symbol $\mathbb{Q}$ represents quantization-friendly layers. 
Llama-3.1-8B is configured with $ \mathbb{Q}=\{0\}$, while Llama-2-7B, Yi-6B, and Yi-9B are configured with $\mathbb{Q}=\{0,1\}$.
Additional details are in \autoref{appendix:experiment_details}.

\paragraph{Hardware.}

The experiments are conducted under two different settings: the first equipped with an NVIDIA RTX 3090 GPU (24GB) and Intel Xeon Gold 6240 CPU, interconnected via PCIe 1.0 ×16 (4GB/s); the second equipped with an NVIDIA A100 GPU (80GB) and Intel Xeon Platinum 8369B CPU, interconnected via PCIe 4.0 ×16 (32GB/s).

\subsection{Accuracy on Long Context Tasks}

\paragraph{LongBench.}

As shown in \autoref{tab:acc}, SnapKV and StreamingLLM degrade in performance due to the loss of crucial information. Although Quest and PQCache improve performance, their individual strategies face limitations under restricted budgets. TailorKV outperforms the best method by 2.32\%, 5.42\%, and 3.66\% on Llama-3.1-8B, Yi-9B, and Yi-6B, respectively, by preserving the 1-bit KV cache for quantization-friendly layers and selecting 192 tokens for sparsity-friendly layers. 
\autoref{appendix:effectiveness_retrieval} provides a discussion on retrieval accuracy of our sparsity-friendly layers compared to other methods.

\paragraph{InfiniteBench.}

\begin{figure}[t]
  \centering
  \includegraphics[width=1\linewidth]{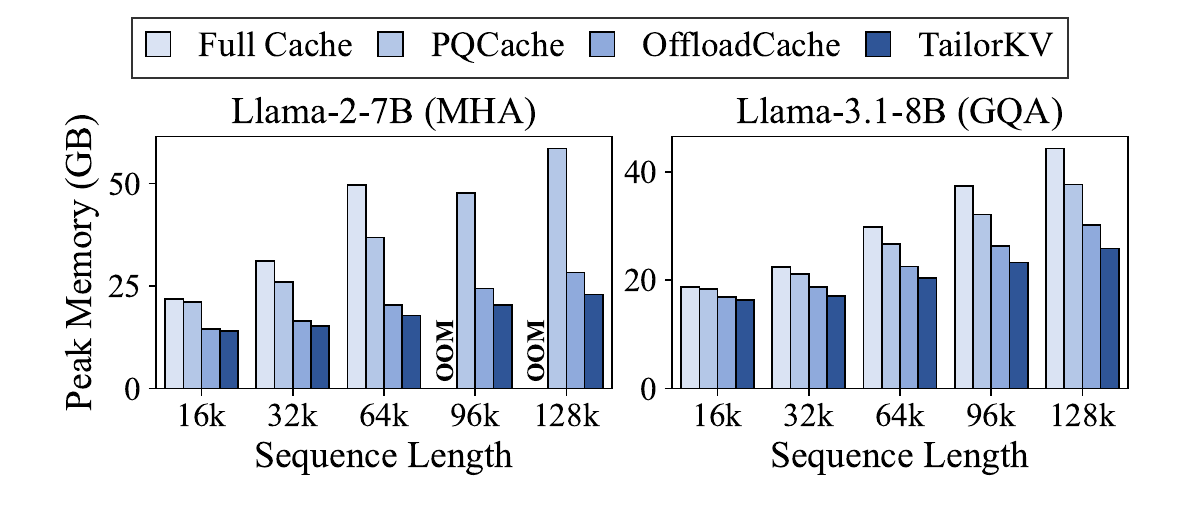} 
\caption{Peak memory usage on A100 (80GB). }
  \label{fig: memory}
\end{figure}
\begin{table}[t]
\centering
\setlength{\tabcolsep}{2.5pt}
\renewcommand{\arraystretch}{1}
\begin{adjustbox}{width=1\linewidth}
    \begin{tabular}{@{}l|cccc|ccc@{}}
    
    \toprule
    \multirow{2}*{\textbf{Methods}} & \multicolumn{4}{c}{\textbf{Llama-2-7B } } &  \multicolumn{3}{c}{\textbf{Llama-3.1-8B } }\\
    \cmidrule(l){2-5}\cmidrule(l){6-8}
   & \textbf{16k} & \textbf{32k} & \textbf{64k} & \textbf{96k} & \textbf{16k} & \textbf{32k} & \textbf{64k} \\

    \midrule
    \multicolumn{8}{c}{NVIDIA RTX 3090 (24GB, PCIe 1.0 link)}
    \\
    \arrayrulecolor{black!30}\midrule
    
     Full Cache & OOM & OOM & OOM & OOM & 0.033 & 0.042 & OOM  \\    
    OffloadCache  & 0.893 & 1.776 & OOM & OOM & 0.242 & 0.460 & OOM   \\
     PQCache  & OOM & OOM & OOM & OOM & 0.126 & OOM & OOM    \\
    TailorKV  & 0.067 & 0.087 & 0.135 & 0.176 & 0.062 & 0.067 & 0.074  \\
    \arrayrulecolor{black} 
    \midrule
    \multicolumn{8}{c}{NVIDIA A100 (80GB, PCIe 4.0 link)}
    \\
    \arrayrulecolor{black!30}\midrule
    Full Cache  & 0.045 & 0.077 & 0.140 & OOM & 0.024 & 0.033 & 0.050   \\
    OffloadCache & 0.433 & 0.838  & 1.767 & 3.253 & 0.124 & 0.227 & 0.435 \\
    PQCache & 0.108 & 0.111 & 0.114 & 0.115  & 0.104 & 0.105& 0.108 \\
     TailorKV  & 0.041 & 0.062 & 0.098 & 0.132 & 0.045 & 0.047 & 0.054 \\
    \arrayrulecolor{black}\bottomrule
    
    \end{tabular}
\end{adjustbox}
  \caption{Decoding latency(s) on different devices. Additional results are provided in \autoref{tab:detail_latency}.} 

  \label{tab: latency}
\end{table}
\autoref{tab:acc} presents evaluations on the challenging benchmark InfiniteBench.
As the context length increases, the clustering overhead of PQCache on the CPU grows. We restrict K-Means to one iteration for real-time inference, which compromises accuracy and exposes PQCache's limitations.
Notably, our hybrid strategy outperforms individual strategies, with an average performance loss under 1.5\% compared to the full cache, especially excelling in dialogue, novel, and math tasks.

\paragraph{RULER.}

\autoref{fig:ruler} summarizes the accuracy on RULER, with the sequence length ranging from 4K to 128K. TailorKV captures crucial information from redundant contexts, leading to superior performance on most tasks, such as Needle-in-a-haystack, Question Answering, and Variable Tracking (detailed results provided in \autoref{tab:appendix_ruler}).

\subsection{Efficiency Results}

We evaluate peak memory usage and decoding latency in comparison with the full cache, OffloadCache, and PQCache. Specifically, the full cache is implemented by FlashAttention-2~\cite{dao2024flashattention} and OffloadCache is a script\footnote{\url{https://github.com/huggingface/transformers}} from the official library that prefetches next layer's KV cache from the CPU memory to the GPU. 
\begin{figure}[t]
  \centering
  \includegraphics[width=1\linewidth]{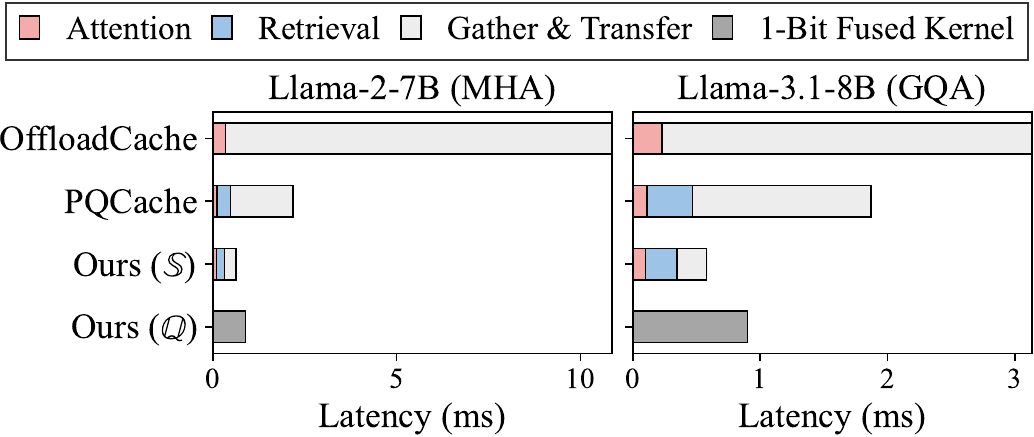} 
\caption{Latency breakdown (ms) under different methods. $\mathbb{Q}$ and $\mathbb{S}$ denote the quantization-friendly layer and sparsity-friendly layer, respectively.}
  \label{fig: break}
\end{figure}

\paragraph{Peak Memory Usage.}

As shown in \autoref{fig: memory}, our method achieves superior memory efficiency compared to alternative methods, enabling deployment on lower-end GPUs such as the RTX 3090. Specifically, compared to full cache, TailorKV reduces GPU memory usage by approximately \(\mathbf{73.8\%}\) for Llama-2-7B with a sequence length of 128k.

\begin{figure*}[t!]
  \centering
  \includegraphics[width=1\linewidth]{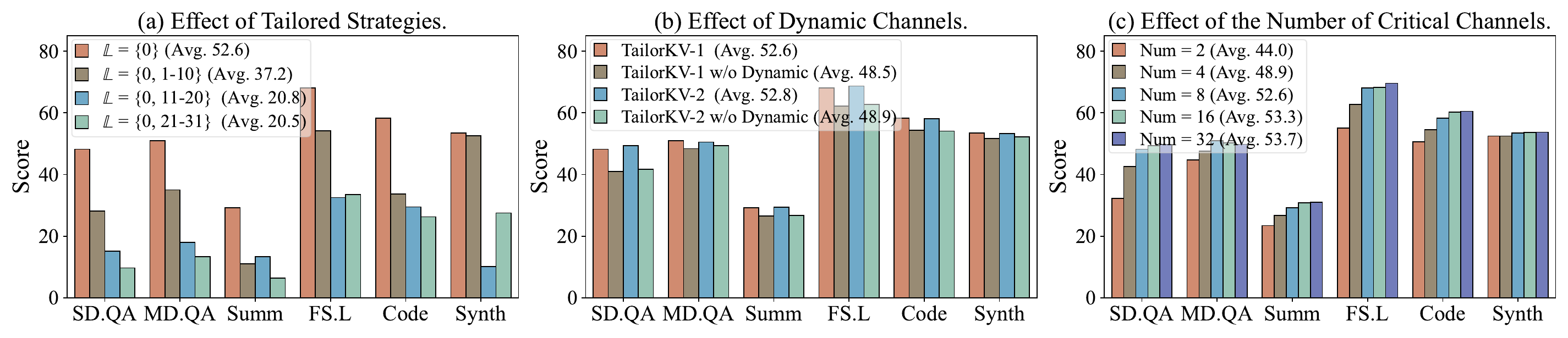} 
\caption{\textbf{Ablation studies.} (a) Performance comparison with different layers quantized to 1-bit. (b) Performance of TailorKV with dynamic or static channels. (c) Performance comparison with different numbers of critical channels. }
  \label{fig: ablation}
\end{figure*}
\paragraph{End-to-End Latency.}
As shown in \autoref{tab: latency}, the increasing sequence length causes out-of-memory errors in the full cache, PQCache, and OffloadCache on the RTX 3090.
For 64k context on the A100, TailorKV achieves significant latency reductions compared to OffloadCache and PQCache: \(\mathbf{18.0}\times\) and \(\mathbf{1.2}\times\) faster than the MHA model, and \(\mathbf{8.1}\times\) and \(\mathbf{2.0}\times\) faster than the GQA model.
TailorKV's latency is comparable to that of full attention, as a result of multi-threading used to execute asynchronous tasks, which enables the overlap of computation and CPU-GPU communication.

\paragraph{Latency Breakdown.}

As depicted in \autoref{fig: break}, we evaluate the breakdown of latency for a single Transformer block with a sequence length of 16k on the A100 GPU.
Compared to PQCache, TailorKV reduces retrieval latency by 27.8\% for the GQA model and 40.5\% for the MHA model, and data transfer latency by 83.5\% and 82.2\% for the same models.
This reduction is primarily attributable to our use of DGL~\cite{wang2019deep} to directly transfer rows from a CPU tensor to the GPU device, whereas PQCache first gathers rows on the CPU and then transfers them to the GPU.

\subsection{Ablation Study}

We conduct ablation studies on the LongBench benchmark using the Llama-3.1-8B-Instruct model.

\paragraph{Effect of Tailored Strategies.}
As depicted in \autoref{fig: ablation} (a), 
1-bit quantization is applied to certain layers, while only 64(+128) tokens are computed for the remaining layers, with the quantization-friendly layer defined as $\mathbb{Q} = \{0\}$. The results indicate that quantizing only the 0th layer yields the best performance, while quantizing sparsity-friendly layers degrades performance, highlighting the need for tailored compression strategies.

\paragraph{Effect of Dynamic Channels.}
\label{sec:ablation}
Prior study~\cite{yang2024post} employed offline calibration to statically select high-magnitude channels. However, we find that outliers may appear at any position, not fixed to specific channels (Section~\ref{insight:channel_outliers}). \autoref{fig: ablation} (b) compares the performance of dynamic and static channels. In general, our dynamic retrieval method demonstrates better performance.

\paragraph{Effect of the Number of Critical Channels.}
In \autoref{fig: ablation} (c), we maintain the  64(+128) configuration and adjust the number of critical channels.
Reducing the number of critical channels decreases retrieval latency. However, performance significantly degrades when the number is set to 2 or 4. Overall, selecting 8 critical channels achieves a favorable balance between performance and latency.

\section{Related Work}
Existing KV cache compression methods include eviction, selection, and quantization, with detailed comparisons in \autoref{appendix:comparison}.
Eviction methods reduce KV cache size by evicting most tokens during inference.
StreamingLLM~\cite{xiao2024efficient} identifies `Attention Sinks' by retaining the initial and the most recent tokens. H2O~\cite{zhang2023h2o}, SnapKV~\cite{li2024snapkv}, and Scissorhands~\cite{liu2024scissorhands} estimate token importance based on historical attention scores. However, evicting dominant tokens may degrade accuracy in tasks like `needle-in-the-haystack' and multi-turn dialogues. 

Selection methods are more commonly used in sparse attention scenarios. Quest~\cite{tang2024quest} retains the KV cache and utilizes paged keys for retrieving tokens, but it fails to reduce memory usage and suffers from lower recall. Instead, KV cache offloading methods like PQCache~\cite{zhang2024pqcache} and InfiniGen~\cite{lee2024infinigen} approximate attention scores for identifying and loading critical tokens from CPU to GPU, though they face challenges in balancing computation and communication due to large KV cache loads. Some methods~\cite{chen2025magicpig,liu2024retrievalattention} use LSH and KNN to retrieve critical tokens, which are processed on the CPU and subsequently merged with GPU outputs; however, imbalanced computation times may result in GPU idle time.

Quantization is a common compression technique that converts high-precision floats into low-precision integers. Existing methods employ various solutions to minimize quantization error. For example, KVQuant~\cite{hooper2024kvquant} isolates outliers for mixed precision, GEAR~\cite{kang2024gear} utilizes SVD to recover residuals, and KIVI~\cite{liu2024kivi} quantizes keys per channel and values per token. FlexGen~\cite{sheng2023flexgen} reduces I/O transfer latency by quantizing the KV cache to 4-bits. However, none of these methods reduce the KV cache to 1-bit. By contrast, we focus on exploring layer characteristics and selecting the most suitable compression strategy.

\section{Conclusion}

In this paper, we propose TailorKV, an effective framework for KV cache management in LLMs. We begin by observing that different layers exhibit distinct compression preferences and categorize them into quantization-friendly and sparsity-friendly, each employing a tailored strategy.
Specifically, quantization-friendly layers aggressively quantize the KV cache to 1-bit. 
Sparsity-friendly layers, on the other hand, dynamically retrieve dominant tokens based on large magnitudes in the query and key channels, integrating CPU-GPU co-design. Experiments across long-context benchmarks show that TailorKV effectively minimizes the usage of the KV cache while maintaining model performance, with an acceptable latency cost.
Our hybrid framework demonstrates the potential of deploying LLMs on resource-limited GPUs, extending the application of LLMs to more devices while maintaining efficiency.

\section*{Limitations}

Although TailorKV has demonstrated superior memory optimization and latency reduction in long-context scenarios, it still exhibits some limitations, which are summarized as follows:
(1) TailorKV primarily focuses on improving the efficiency of the decode phase by asynchronously transferring tokens from the CPU memory to the GPU. However, it is challenging to completely overlap the offloading latency during the prefill phase. Moreover, the efficiency of the prefill phase in long-context scenarios is also important. It is noteworthy that our work is compatible with and complementary to approaches for prefilling acceleration~\cite{jiang2024minference}.
(2) We have designed tailored strategies for different layers to facilitate deployment, and we are confident that TailorKV can be adapted on a head-wise basis. 
These issues hold significant importance, and we intend to further explore them in our future research.

\section*{Acknowledgments}

This work was supported by the National Natural Science Foundation of China (Nos. 62472419, 62472420). We would like to thank the anonymous reviewers for their insightful comments.

\bibliography{custom}

\begin{thebibliography}{40}
\providecommand{\natexlab}[1]{#1}

\bibitem[{{01-ai}(2024{\natexlab{a}})}]{yi6b}
{01-ai}. 2024{\natexlab{a}}.
\newblock Yi-6b-200k.
\newblock \url{https://huggingface.co/01-ai/Yi-6B-200K}.
\newblock Accessed: 2024-07-01.

\bibitem[{{01-ai}(2024{\natexlab{b}})}]{yi9b}
{01-ai}. 2024{\natexlab{b}}.
\newblock Yi-9b-200k.
\newblock \url{https://huggingface.co/01-ai/Yi-9B-200K}.
\newblock Accessed: 2024-07-01.

\bibitem[{Achiam et~al.(2023)Achiam, Adler, Agarwal, Ahmad, Akkaya, Aleman, Almeida, Altenschmidt, Altman, Anadkat et~al.}]{achiam2023gpt}
Josh Achiam, Steven Adler, Sandhini Agarwal, Lama Ahmad, Ilge Akkaya, Florencia~Leoni Aleman, Diogo Almeida, Janko Altenschmidt, Sam Altman, Shyamal Anadkat, et~al. 2023.
\newblock \href {https://arxiv.org/abs/2303.08774} {Gpt-4 technical report}.
\newblock \emph{ArXiv preprint}, abs/2303.08774.

\bibitem[{Bai et~al.(2024)Bai, Lv, Zhang, Lyu, Tang, Huang, Du, Liu, Zeng, Hou, Dong, Tang, and Li}]{bai-etal-2024-longbench}
Yushi Bai, Xin Lv, Jiajie Zhang, Hongchang Lyu, Jiankai Tang, Zhidian Huang, Zhengxiao Du, Xiao Liu, Aohan Zeng, Lei Hou, Yuxiao Dong, Jie Tang, and Juanzi Li. 2024.
\newblock \href {https://doi.org/10.18653/v1/2024.acl-long.172} {{L}ong{B}ench: A bilingual, multitask benchmark for long context understanding}.
\newblock In \emph{Proc. of ACL}, pages 3119--3137. Association for Computational Linguistics.

\bibitem[{Cai et~al.(2024)Cai, Zhang, Gao, Liu, Liu, Lu, Xiong, Dong, Chang, Hu et~al.}]{cai2024pyramidkv}
Zefan Cai, Yichi Zhang, Bofei Gao, Yuliang Liu, Tianyu Liu, Keming Lu, Wayne Xiong, Yue Dong, Baobao Chang, Junjie Hu, et~al. 2024.
\newblock \href {https://arxiv.org/abs/2406.02069} {Pyramidkv: Dynamic kv cache compression based on pyramidal information funneling}.
\newblock \emph{ArXiv preprint}, abs/2406.02069.

\bibitem[{Chen et~al.(2025)Chen, Sadhukhan, Ye, Zhou, Zhang, Nolte, Tian, Douze, Bottou, Jia, and Chen}]{chen2025magicpig}
Zhuoming Chen, Ranajoy Sadhukhan, Zihao Ye, Yang Zhou, Jianyu Zhang, Niklas Nolte, Yuandong Tian, Matthijs Douze, Leon Bottou, Zhihao Jia, and Beidi Chen. 2025.
\newblock \href {https://openreview.net/forum?id=ALzTQUgW8a} {Magic{PIG}: {LSH} sampling for efficient {LLM} generation}.
\newblock In \emph{The Thirteenth International Conference on Learning Representations}.

\bibitem[{Chiang et~al.(2023)Chiang, Li, Lin, Sheng, Wu, Zhang, Zheng, Zhuang, Zhuang, Gonzalez et~al.}]{chiang2023vicuna}
Wei-Lin Chiang, Zhuohan Li, Zi~Lin, Ying Sheng, Zhanghao Wu, Hao Zhang, Lianmin Zheng, Siyuan Zhuang, Yonghao Zhuang, Joseph~E Gonzalez, et~al. 2023.
\newblock Vicuna: An open-source chatbot impressing gpt-4 with 90\%* chatgpt quality.
\newblock \emph{See https://vicuna. lmsys. org (accessed 14 April 2023)}, 2(3):6.

\bibitem[{Dao(2024)}]{dao2024flashattention}
Tri Dao. 2024.
\newblock \href {https://openreview.net/forum?id=mZn2Xyh9Ec} {Flashattention-2: Faster attention with better parallelism and work partitioning}.
\newblock In \emph{The Twelfth International Conference on Learning Representations}.

\bibitem[{Dubey et~al.(2024)Dubey, Jauhri, Pandey, Kadian, Al-Dahle, Letman, Mathur, Schelten, Yang, Fan et~al.}]{dubey2024llama}
Abhimanyu Dubey, Abhinav Jauhri, Abhinav Pandey, Abhishek Kadian, Ahmad Al-Dahle, Aiesha Letman, Akhil Mathur, Alan Schelten, Amy Yang, Angela Fan, et~al. 2024.
\newblock \href {https://arxiv.org/abs/2407.21783} {The llama 3 herd of models}.
\newblock \emph{ArXiv preprint}, abs/2407.21783.

\bibitem[{Feng et~al.(2024)Feng, Lv, Cao, Xie, and Zhou}]{feng2024ada}
Yuan Feng, Junlin Lv, Yukun Cao, Xike Xie, and S~Kevin Zhou. 2024.
\newblock \href {https://arxiv.org/abs/2407.11550} {Ada-kv: Optimizing kv cache eviction by adaptive budget allocation for efficient llm inference}.
\newblock \emph{ArXiv preprint}, abs/2407.11550.

\bibitem[{He et~al.(2024)He, Zhang, Wu, Liu, Zhou, and Zhuang}]{he2024zipcache}
Yefei He, Luoming Zhang, Weijia Wu, Jing Liu, Hong Zhou, and Bohan Zhuang. 2024.
\newblock \href {https://openreview.net/forum?id=5t4ZAkPiJs} {Zipcache: Accurate and efficient {KV} cache quantization with salient token identification}.
\newblock In \emph{The Thirty-eighth Annual Conference on Neural Information Processing Systems}.

\bibitem[{Hooper et~al.(2024)Hooper, Kim, Mohammadzadeh, Mahoney, Shao, Keutzer, and Gholami}]{hooper2024kvquant}
Coleman Hooper, Sehoon Kim, Hiva Mohammadzadeh, Michael~W Mahoney, Yakun~S Shao, Kurt Keutzer, and Amir Gholami. 2024.
\newblock Kvquant: Towards 10 million context length llm inference with kv cache quantization.
\newblock \emph{Advances in Neural Information Processing Systems}, 37:1270--1303.

\bibitem[{Hsieh et~al.(2024)Hsieh, Sun, Kriman, Acharya, Rekesh, Jia, and Ginsburg}]{hsieh2024ruler}
Cheng-Ping Hsieh, Simeng Sun, Samuel Kriman, Shantanu Acharya, Dima Rekesh, Fei Jia, and Boris Ginsburg. 2024.
\newblock \href {https://openreview.net/forum?id=kIoBbc76Sy} {{RULER}: What{\textquoteright}s the real context size of your long-context language models?}
\newblock In \emph{First Conference on Language Modeling}.

\bibitem[{Jiang et~al.(2024)Jiang, Li, Zhang, Wu, Luo, Ahn, Han, Abdi, Li, Lin, Yang, and Qiu}]{jiang2024minference}
Huiqiang Jiang, Yucheng Li, Chengruidong Zhang, Qianhui Wu, Xufang Luo, Surin Ahn, Zhenhua Han, Amir~H. Abdi, Dongsheng Li, Chin-Yew Lin, Yuqing Yang, and Lili Qiu. 2024.
\newblock \href {https://openreview.net/forum?id=fPBACAbqSN} {{MI}nference 1.0: Accelerating pre-filling for long-context {LLM}s via dynamic sparse attention}.
\newblock In \emph{The Thirty-eighth Annual Conference on Neural Information Processing Systems}.

\bibitem[{Kang et~al.(2024)Kang, Zhang, Kundu, Jeong, Liu, Krishna, and Zhao}]{kang2024gear}
Hao Kang, Qingru Zhang, Souvik Kundu, Geonhwa Jeong, Zaoxing Liu, Tushar Krishna, and Tuo Zhao. 2024.
\newblock \href {https://arxiv.org/abs/2403.05527} {Gear: An efficient kv cache compression recipefor near-lossless generative inference of llm}.
\newblock \emph{ArXiv preprint}, abs/2403.05527.

\bibitem[{Kwon et~al.(2023)Kwon, Li, Zhuang, Sheng, Zheng, Yu, Gonzalez, Zhang, and Stoica}]{kwon2023efficient}
Woosuk Kwon, Zhuohan Li, Siyuan Zhuang, Ying Sheng, Lianmin Zheng, Cody~Hao Yu, Joseph Gonzalez, Hao Zhang, and Ion Stoica. 2023.
\newblock Efficient memory management for large language model serving with pagedattention.
\newblock In \emph{Proceedings of the 29th Symposium on Operating Systems Principles}, pages 611--626.

\bibitem[{Lee et~al.(2024)Lee, Lee, Seo, and Sim}]{lee2024infinigen}
Wonbeom Lee, Jungi Lee, Junghwan Seo, and Jaewoong Sim. 2024.
\newblock Infinigen: Efficient generative inference of large language models with dynamic kv cache management.
\newblock In \emph{18th USENIX Symposium on Operating Systems Design and Implementation (OSDI 24)}, pages 155--172.

\bibitem[{Li et~al.(2024)Li, Huang, Yang, Venkitesh, Locatelli, Ye, Cai, Lewis, and Chen}]{li2024snapkv}
Yuhong Li, Yingbing Huang, Bowen Yang, Bharat Venkitesh, Acyr Locatelli, Hanchen Ye, Tianle Cai, Patrick Lewis, and Deming Chen. 2024.
\newblock \href {https://openreview.net/forum?id=poE54GOq2l} {Snap{KV}: {LLM} knows what you are looking for before generation}.
\newblock In \emph{The Thirty-eighth Annual Conference on Neural Information Processing Systems}.

\bibitem[{Lin et~al.(2024)Lin, Tang, Tang, Yang, Chen, Wang, Xiao, Dang, Gan, and Han}]{lin2024awq}
Ji~Lin, Jiaming Tang, Haotian Tang, Shang Yang, Wei-Ming Chen, Wei-Chen Wang, Guangxuan Xiao, Xingyu Dang, Chuang Gan, and Song Han. 2024.
\newblock Awq: Activation-aware weight quantization for on-device llm compression and acceleration.
\newblock \emph{Proceedings of Machine Learning and Systems}, 6:87--100.

\bibitem[{Liu et~al.(2024{\natexlab{a}})Liu, Feng, Xue, Wang, Wu, Lu, Zhao, Deng, Zhang, Ruan et~al.}]{liu2024deepseek}
Aixin Liu, Bei Feng, Bing Xue, Bingxuan Wang, Bochao Wu, Chengda Lu, Chenggang Zhao, Chengqi Deng, Chenyu Zhang, Chong Ruan, et~al. 2024{\natexlab{a}}.
\newblock Deepseek-v3 technical report.
\newblock \emph{arXiv preprint arXiv:2412.19437}.

\bibitem[{Liu et~al.(2024{\natexlab{b}})Liu, Liu, Pan, He, Haffari, and Zhuang}]{liu2024minicache}
Akide Liu, Jing Liu, Zizheng Pan, Yefei He, Reza Haffari, and Bohan Zhuang. 2024{\natexlab{b}}.
\newblock Minicache: Kv cache compression in depth dimension for large language models.
\newblock \emph{Advances in Neural Information Processing Systems}, 37:139997--140031.

\bibitem[{Liu et~al.(2024{\natexlab{c}})Liu, Chen, Lu, Jiang, Han, Zhang, Chen, Zhang, Ding, Zhang et~al.}]{liu2024retrievalattention}
Di~Liu, Meng Chen, Baotong Lu, Huiqiang Jiang, Zhenhua Han, Qianxi Zhang, Qi~Chen, Chengruidong Zhang, Bailu Ding, Kai Zhang, et~al. 2024{\natexlab{c}}.
\newblock \href {https://arxiv.org/abs/2409.10516} {Retrievalattention: Accelerating long-context llm inference via vector retrieval}.
\newblock \emph{ArXiv preprint}, abs/2409.10516.

\bibitem[{Liu et~al.(2023)Liu, Desai, Liao, Wang, Xie, Xu, Kyrillidis, and Shrivastava}]{liu2024scissorhands}
Zichang Liu, Aditya Desai, Fangshuo Liao, Weitao Wang, Victor Xie, Zhaozhuo Xu, Anastasios Kyrillidis, and Anshumali Shrivastava. 2023.
\newblock \href {http://papers.nips.cc/paper\_files/paper/2023/hash/a452a7c6c463e4ae8fbdc614c6e983e6-Abstract-Conference.html} {Scissorhands: Exploiting the persistence of importance hypothesis for {LLM} {KV} cache compression at test time}.
\newblock In \emph{Proc. of NeurIPS}.

\bibitem[{Liu et~al.(2024{\natexlab{d}})Liu, Yuan, Jin, Zhong, Xu, Braverman, Chen, and Hu}]{liu2024kivi}
Zirui Liu, Jiayi Yuan, Hongye Jin, Shaochen~(Henry) Zhong, Zhaozhuo Xu, Vladimir Braverman, Beidi Chen, and Xia Hu. 2024{\natexlab{d}}.
\newblock Kivi: a tuning-free asymmetric 2bit quantization for kv cache.
\newblock In \emph{Proceedings of the 41st International Conference on Machine Learning}, ICML'24. JMLR.org.

\bibitem[{Qin et~al.(2025)Qin, Li, He, Cui, Ren, Zhang, Wu, Zheng, and Xu}]{qin2025mooncake}
Ruoyu Qin, Zheming Li, Weiran He, Jialei Cui, Feng Ren, Mingxing Zhang, Yongwei Wu, Weimin Zheng, and Xinran Xu. 2025.
\newblock \href {https://www.usenix.org/conference/fast25/presentation/qin} {Mooncake: Trading more storage for less computation {\textemdash} a {KVCache-centric} architecture for serving {LLM} chatbot}.
\newblock In \emph{23rd USENIX Conference on File and Storage Technologies (FAST 25)}, pages 155--170, Santa Clara, CA. USENIX Association.

\bibitem[{Ribar et~al.(2024)Ribar, Chelombiev, Hudlass-Galley, Blake, Luschi, and Orr}]{ribar2024sparq}
Luka Ribar, Ivan Chelombiev, Luke Hudlass-Galley, Charlie Blake, Carlo Luschi, and Douglas Orr. 2024.
\newblock \href {https://openreview.net/forum?id=Ue8EHzaFI4} {Sparq attention: Bandwidth-efficient {LLM} inference}.
\newblock In \emph{ICLR 2024 Workshop on Mathematical and Empirical Understanding of Foundation Models}.

\bibitem[{Sheng et~al.(2023)Sheng, Zheng, Yuan, Li, Ryabinin, Chen, Liang, R{\'{e}}, Stoica, and Zhang}]{sheng2023flexgen}
Ying Sheng, Lianmin Zheng, Binhang Yuan, Zhuohan Li, Max Ryabinin, Beidi Chen, Percy Liang, Christopher R{\'{e}}, Ion Stoica, and Ce~Zhang. 2023.
\newblock \href {https://proceedings.mlr.press/v202/sheng23a.html} {Flexgen: High-throughput generative inference of large language models with a single {GPU}}.
\newblock In \emph{Proc. of ICML}, volume 202 of \emph{Proceedings of Machine Learning Research}, pages 31094--31116. {PMLR}.

\bibitem[{Tang et~al.(2024)Tang, Zhao, Zhu, Xiao, Kasikci, and Han}]{tang2024quest}
Jiaming Tang, Yilong Zhao, Kan Zhu, Guangxuan Xiao, Baris Kasikci, and Song Han. 2024.
\newblock Quest: query-aware sparsity for efficient long-context llm inference.
\newblock In \emph{Proceedings of the 41st International Conference on Machine Learning}, ICML'24. JMLR.org.

\bibitem[{Touvron et~al.(2023)Touvron, Lavril, Izacard, Martinet, Lachaux, Lacroix, Rozi{\`e}re, Goyal, Hambro, Azhar et~al.}]{touvron2023llama}
Hugo Touvron, Thibaut Lavril, Gautier Izacard, Xavier Martinet, Marie-Anne Lachaux, Timoth{\'e}e Lacroix, Baptiste Rozi{\`e}re, Naman Goyal, Eric Hambro, Faisal Azhar, et~al. 2023.
\newblock Llama: Open and efficient foundation language models.
\newblock \emph{arXiv preprint arXiv:2302.13971}.

\bibitem[{Wan et~al.(2025)Wan, Wu, Zhang, Xin, Tao, Zhu, Wang, Luo, Xiong, Wang, and Zhang}]{wan2025textdtexto}
Zhongwei Wan, Xinjian Wu, Yu~Zhang, Yi~Xin, Chaofan Tao, Zhihong Zhu, Xin Wang, Siqi Luo, Jing Xiong, Longyue Wang, and Mi~Zhang. 2025.
\newblock \href {https://openreview.net/forum?id=HzBfoUdjHt} {\${\textbackslash}text\{D\}\_\{2\}{\textbackslash}text\{O\}\$: Dynamic discriminative operations for efficient long-context inference of large language models}.
\newblock In \emph{The Thirteenth International Conference on Learning Representations}.

\bibitem[{Wang et~al.(2019)Wang, Zheng, Ye, Gan, Li, Song, Zhou, Ma, Yu, Gai et~al.}]{wang2019deep}
Minjie Wang, Da~Zheng, Zihao Ye, Quan Gan, Mufei Li, Xiang Song, Jinjing Zhou, Chao Ma, Lingfan Yu, Yu~Gai, et~al. 2019.
\newblock \href {https://arxiv.org/abs/1909.01315} {Deep graph library: A graph-centric, highly-performant package for graph neural networks}.
\newblock \emph{ArXiv preprint}, abs/1909.01315.

\bibitem[{Xiao et~al.(2024{\natexlab{a}})Xiao, Zhang, Han, Xiao, Lin, Zhang, Liu, and Sun}]{xiao2024infllm}
Chaojun Xiao, Pengle Zhang, Xu~Han, Guangxuan Xiao, Yankai Lin, Zhengyan Zhang, Zhiyuan Liu, and Maosong Sun. 2024{\natexlab{a}}.
\newblock \href {https://openreview.net/forum?id=bTHFrqhASY} {Inf{LLM}: Training-free long-context extrapolation for {LLM}s with an efficient context memory}.
\newblock In \emph{The Thirty-eighth Annual Conference on Neural Information Processing Systems}.

\bibitem[{Xiao et~al.(2024{\natexlab{b}})Xiao, Tian, Chen, Han, and Lewis}]{xiao2024efficient}
Guangxuan Xiao, Yuandong Tian, Beidi Chen, Song Han, and Mike Lewis. 2024{\natexlab{b}}.
\newblock \href {https://openreview.net/forum?id=NG7sS51zVF} {Efficient streaming language models with attention sinks}.
\newblock In \emph{The Twelfth International Conference on Learning Representations}.

\bibitem[{Yang et~al.(2024{\natexlab{a}})Yang, Kim, Bae, Kwon, Park, Yang, Kwon, and Lee}]{yang2024no}
June~Yong Yang, Byeongwook Kim, Jeongin Bae, Beomseok Kwon, Gunho Park, Eunho Yang, Se~Jung Kwon, and Dongsoo Lee. 2024{\natexlab{a}}.
\newblock \href {https://arxiv.org/abs/2402.18096} {No token left behind: Reliable kv cache compression via importance-aware mixed precision quantization}.
\newblock \emph{ArXiv preprint}, abs/2402.18096.

\bibitem[{Yang et~al.(2024{\natexlab{b}})Yang, Sheng, Gonzalez, Stoica, and Zheng}]{yang2024post}
Shuo Yang, Ying Sheng, Joseph~E Gonzalez, Ion Stoica, and Lianmin Zheng. 2024{\natexlab{b}}.
\newblock \href {https://arxiv.org/abs/2408.07092} {Post-training sparse attention with double sparsity}.
\newblock \emph{ArXiv preprint}, abs/2408.07092.

\bibitem[{Zhang et~al.(2024{\natexlab{a}})Zhang, Ji, Chen, Fu, Miao, Nie, Chen, and Cui}]{zhang2024pqcache}
Hailin Zhang, Xiaodong Ji, Yilin Chen, Fangcheng Fu, Xupeng Miao, Xiaonan Nie, Weipeng Chen, and Bin Cui. 2024{\natexlab{a}}.
\newblock \href {https://arxiv.org/abs/2407.12820} {Pqcache: Product quantization-based kvcache for long context llm inference}.
\newblock \emph{ArXiv preprint}, abs/2407.12820.

\bibitem[{Zhang et~al.(2024{\natexlab{b}})Zhang, Zhang, Xu, Mei, and Li}]{zhang2024dovetail}
Libo Zhang, Zhaoning Zhang, Baizhou Xu, Songzhu Mei, and Dongsheng Li. 2024{\natexlab{b}}.
\newblock Dovetail: A cpu/gpu heterogeneous speculative decoding for llm inference.
\newblock \emph{arXiv preprint arXiv:2412.18934}.

\bibitem[{Zhang et~al.(2024{\natexlab{c}})Zhang, Chen, Hu, Xu, Chen, Hao, Han, Thai, Wang, Liu, and Sun}]{zhang-etal-2024-bench}
Xinrong Zhang, Yingfa Chen, Shengding Hu, Zihang Xu, Junhao Chen, Moo Hao, Xu~Han, Zhen Thai, Shuo Wang, Zhiyuan Liu, and Maosong Sun. 2024{\natexlab{c}}.
\newblock \href {https://doi.org/10.18653/v1/2024.acl-long.814} {$\infty${B}ench: Extending long context evaluation beyond 100{K} tokens}.
\newblock In \emph{Proceedings of the 62nd Annual Meeting of the Association for Computational Linguistics (Volume 1: Long Papers)}, pages 15262--15277, Bangkok, Thailand. Association for Computational Linguistics.

\bibitem[{Zhang et~al.(2024{\natexlab{d}})Zhang, Du, Du, Pang, Gao, and Lin}]{zhang2024simlayerkv}
Xuan Zhang, Cunxiao Du, Chao Du, Tianyu Pang, Wei Gao, and Min Lin. 2024{\natexlab{d}}.
\newblock \href {https://arxiv.org/abs/2410.13846} {Simlayerkv: A simple framework for layer-level kv cache reduction}.
\newblock \emph{ArXiv preprint}, abs/2410.13846.

\bibitem[{Zhang et~al.(2023)Zhang, Sheng, Zhou, Chen, Zheng, Cai, Song, Tian, R{\'{e}}, Barrett, Wang, and Chen}]{zhang2023h2o}
Zhenyu Zhang, Ying Sheng, Tianyi Zhou, Tianlong Chen, Lianmin Zheng, Ruisi Cai, Zhao Song, Yuandong Tian, Christopher R{\'{e}}, Clark~W. Barrett, Zhangyang Wang, and Beidi Chen. 2023.
\newblock \href {http://papers.nips.cc/paper\_files/paper/2023/hash/6ceefa7b15572587b78ecfcebb2827f8-Abstract-Conference.html} {{H2O:} heavy-hitter oracle for efficient generative inference of large language models}.
\newblock In \emph{Proc. of NeurIPS}.

\end{thebibliography}

\newpage

\appendix

\section{Comparison with Other Approaches}
\label{appendix:comparison}

\begin{figure}[thb!]
\centering
  \includegraphics[width=1\linewidth]{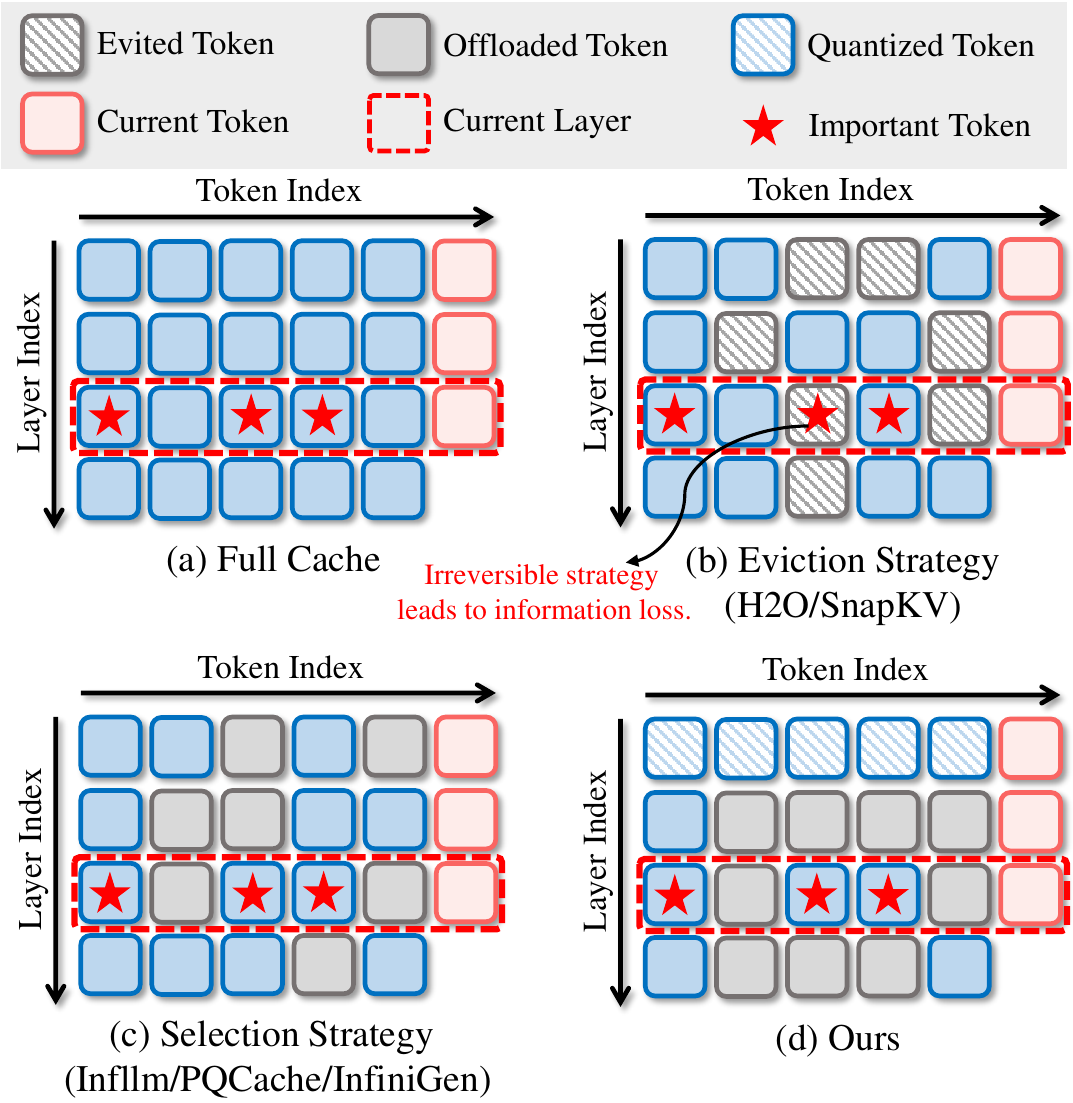} 
\caption{Comparison of TailorKV with other methods in managing KV cache budget across layers.}
    \label{fig:compare}
\end{figure}

\autoref{fig:compare} compares TailorKV with other methods: (a) Full cache retains the entire KV cache. (b) The eviction methods permanently evict specific tokens from each layer, leading to irreversible information loss since evicted tokens may be important later. (c) The selection methods offload the entire KV cache to the CPU, enabling tokens recall but incurring significant communication overhead because of the large volume of tokens involved. (d) Our method employs layer-specific compression strategies, facilitating more aggressive compression.

\section{Inter-Layer Similarity}
\label{appendix:similarity}
Let \( \mathbf{h}^{(l)} \) denote the hidden state at the \( l \)-th layer. To quantify the similarity between the hidden states of two adjacent layers, we employ cosine similarity, which is formally defined as:
\begin{equation}
\mathrm{sim}(\mathbf{h}^{(l-1)}, \mathbf{h}^{(l)}) = \frac{\mathbf{h}^{(l-1)} \cdot \mathbf{h}^{(l)}}{\|\mathbf{h}^{(l-1)}\| \|\mathbf{h}^{(l)}\|}.
\end{equation}
We define the query weight at the \( l \)-th layer as \( W_q^{(l)} \), and the query vector at the \( l \)-th layer is computed as:
\begin{equation}
{\mathbf{q}}^{(l)} = \mathbf{W}_{q}^{(l)}(\mathbf{h}^{(l)}).
\end{equation}

\begin{figure}[thb!]
  \centering
  \includegraphics[width=0.85\linewidth]{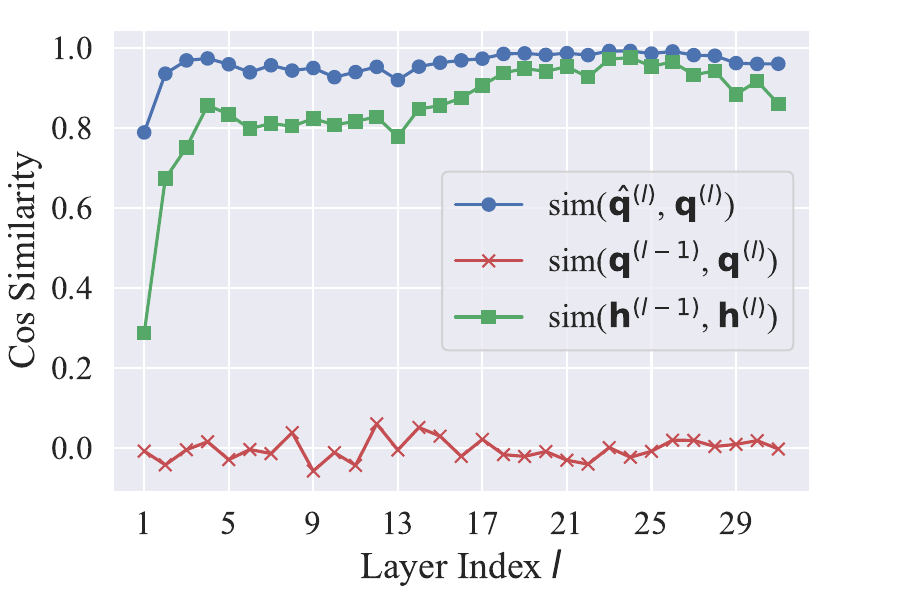} 
  \caption{Cosine similarity between adjacent layers. }

  \label{fig: cos}
\end{figure}

As shown in \autoref{fig: cos}, \( \mathbf{h}^{(l)} \) and \( \mathbf{h}^{(l-1)} \) closely resemble each other, allowing us to approximate the query at the \( l \)-th layer based on the hidden state from the \( l-1 \)-th layer:
\begin{equation}
\hat{\mathbf{q}}^{(l)} = \mathbf{W}_{q}^{(l)}(\mathbf{h}^{(l-1)}).
\end{equation}

Existing research~\cite{liu2024minicache} has elucidated that the KV cache exhibits similarity across adjacent layers. However, as illustrated in \autoref{fig: cos},  the similarity between \(\hat{\mathbf{q}}^{(l)}\) and \(\mathbf{q}^{(l)}\) exceeds that between \(\mathbf{q}^{(l-1)}\) and \(\mathbf{q}^{(l)}\), suggesting that using hidden states from the preceding layer enhances prediction accuracy.

\section{Offline Identification on Different Datasets}
\label{appendix:offline}

\begin{figure}[thb!]
  \centering
  \includegraphics[width=0.75\linewidth]{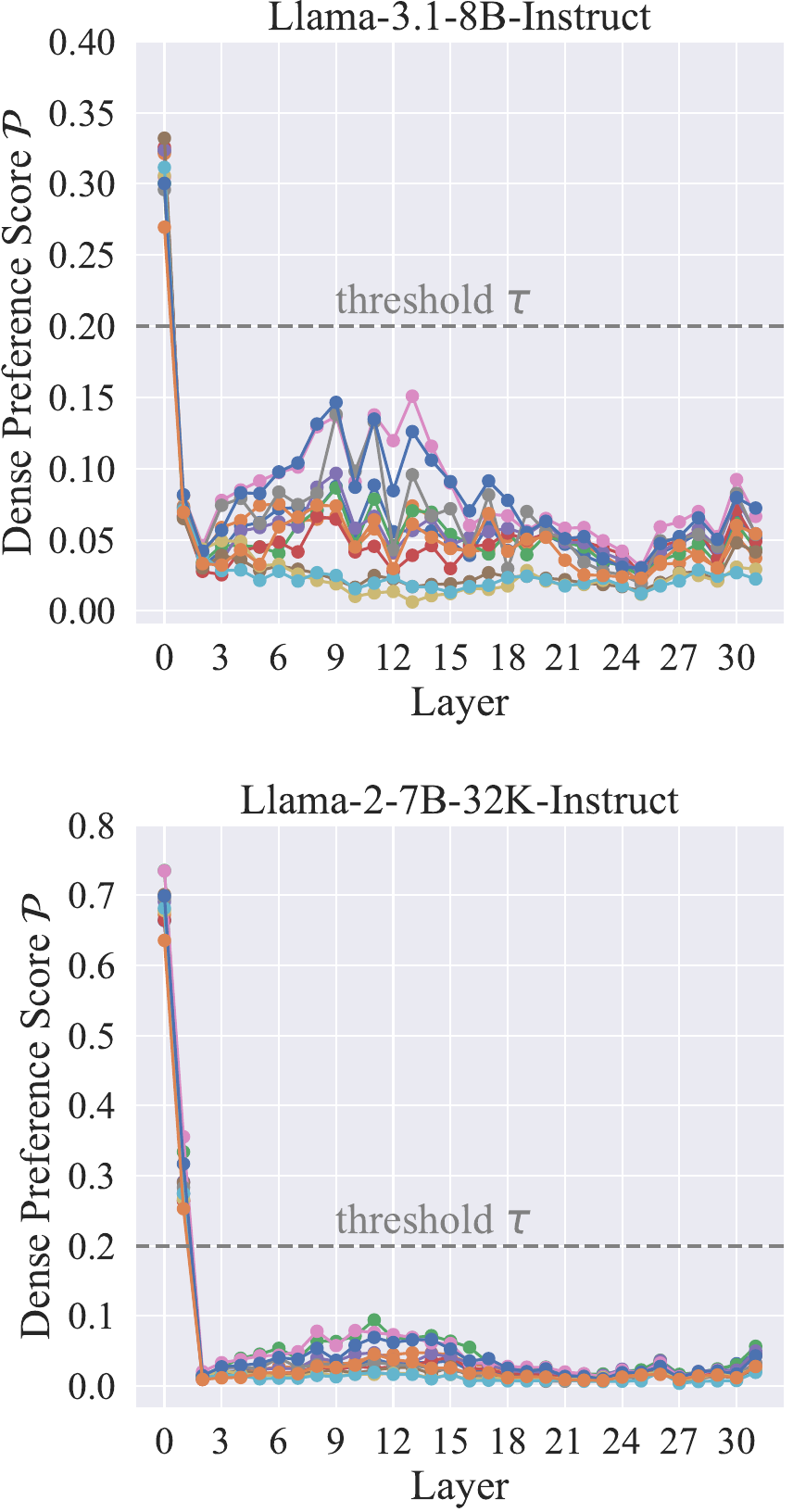} 
  \caption{Dense preference score \( \mathcal{P} \) for layers across different offline datasets.}
  \label{fig: Identification}
\end{figure}

As shown in \autoref{fig: Identification}, the curves represent different datasets. The distribution of \( \mathcal{P} \) is consistent across various datasets for the same model, indicating that the metric \( \mathcal{P} \) effectively captures the characteristics of different layers, enabling appropriate compression strategy.

\section{Baselines Settings}
\label{appendix:experiment_details}

In \autoref{tab:experiment_details1} and \autoref{tab:experiment_details2}, we present the configuration for the long-context methods employed in our experiments.
\begin{table}[thb!]
\centering
\small
\setlength{\tabcolsep}{4pt}
\begin{tabular}{@{}lp{5.2cm}@{}}

\toprule
\textbf{Methods} & \textbf{Configurations}  \\
\midrule
StreamingLLM & num local: 128, num initial: 64\\
\midrule
SnapKV & window size: 64, max capacity prompt: 128, kernel size: 7, pooling: max pooling\\
\midrule
Quest &  page size: 16, token budget: 196 \\
\midrule
PQCache & partitions in PQ: 2, bits for PQ codes: 6, K-Means iterations: adaptive, \(n_{\text{local}}\): 64, \(n_{\text{topk}}\): 128\\
\midrule
\texttt{TailorKV-1} &\( \tau \): 0.2, bit size: 1, group size: 64, \(n_{\text{local}}\): 64, \(n_{\text{topk}}\): 128,  num critical channels: 8\\
\midrule
\texttt{TailorKV-2} &\( \tau \): 0.2, bit size: 2, group size: 64, \(n_{\text{local}}\): 64, \(n_{\text{topk}}\): 128,  num critical channels: 8\\
\bottomrule
\end{tabular}
  \caption{Configurations of long-context methods on \textbf{LongBench}.}
  \label{tab:experiment_details1}
\end{table}

\begin{table}[thb!]
\centering
\small
\setlength{\tabcolsep}{4pt}
\begin{tabular}{@{}lp{5.2cm}@{}}

\toprule
\textbf{Methods} & \textbf{Configurations}  \\
\midrule
StreamingLLM & num local: 896, num initial: 128\\
\midrule
SnapKV & window size: 128, max capacity prompt: 896, kernel size: 7, pooling: max pooling\\
\midrule
Quest &  page size: 16, token budget: 1024\\
\midrule
PQCache & partitions in PQ: 2, bits for PQ codes: 6, K-Means iterations: 1 (exceeding 64k), \(n_{\text{local}}\): 128, \(n_{\text{topk}}\): 896\\
\midrule
\texttt{TailorKV-1} &\( \tau \): 0.2, bit size: 1, group size: 64, \(n_{\text{local}}\): 128, \(n_{\text{topk}}\): 896, num critical channels: 12\\
\midrule
\texttt{TailorKV-2} &\( \tau \): 0.2, bit size: 2, group size: 64, \(n_{\text{local}}\): 128, \(n_{\text{topk}}\): 896, num critical channels: 12\\
\bottomrule
\end{tabular}
  \caption{Configurations of long-context methods on \textbf{InfiniteBench} and \textbf{RULER}.}
  \label{tab:experiment_details2}
\end{table}

\section{Comparison with Hybrid Method}
\begin{table*}[t]
\centering
\small
\setlength{\tabcolsep}{1.7pt}

\begin{tabular}{lccccccccccccccc}
\toprule
\textbf{Methos} & \textbf{Ratio} & \textbf{Qspr} & \textbf{MulFi} & \textbf{HQA} & \textbf{WMQA} & \textbf{GRpt} & \textbf{MulN} & \textbf{TREC} & \textbf{SMSM} & \textbf{TriQA} & \textbf{Repo} & \textbf{LCC } & \textbf{PsgC} & \textbf{PsgR} & \textbf{Avg.}\\
\midrule
\textit{Llama-3.1-8B} & \(1 \times\) & 45.5 &53.8 & 54.7 &47.1 & 34.9 &27.5 &73.0 & 43.8 &91.6 & 56.7 & 63.4 & 7.5 & 99.5 & 53.8 \\
\midrule
SimLayerKV& \(1.53 \times\) & \textbf{45.6} & 52.3 & 54.5 & 44.5 & \textbf{32.2} & 26.9 & \textbf{71.5} & \textbf{43.8} & 91.3 & \textbf{54.9}& \textbf{62.8} & \textbf{7.9} & 95.5 & 52.6 \\
\texttt{TailorKV-1} & \(34.2 \times\) & 43.4 & 53.0 & \textbf{55.3} & \textbf{46.5} & 31.3 & \textbf{27.2} & 70.0 & 42.5 & 91.6 & 54.8 & 61.8 & \textbf{7.9} & \textbf{99.0} & 52.6 \\
\texttt{TailorKV-2} & \(32.7 \times\) & 44.8 & \textbf{53.9} & 54.8 & 46.2 & 31.9 & 26.8 & 70.5 & 43.2 & \textbf{92.3} & 54.2 & 62.1 & 7.7 & \textbf{99.0} & \textbf{52.9}\\

\bottomrule
\end{tabular}
\caption{Performance comparison between TailorKV and SimLayerKV. TailorKV computes only 64 (+128) tokens for sparsity-friendly layers. SimLayerKV retains the most recent 1024 tokens for the "lazy" layers, while the "non-lazy" layers preserve full precision. Additionally, the threshold for SimLayerKV on Llama-3.1-8B is 0.9, with more than half of the layers being "non-lazy."}
\label{tab:simlayer}
\end{table*}
To validate the effectiveness of our quantization-sparsity hybrid framework, we compare it to SimLayerKV~\cite{zhang2024simlayerkv}, a similar hybrid method. SimLayerKV assumes that some layers in LLMs exhibit "lazy" behavior, retaining only the initial and most recent tokens, while "non-lazy" layers require full precision to retain all tokens.
\autoref{tab:simlayer} presents the experimental results on LongBench. The results show that at an average compression rate of \(34.2 \times \), the performance of our method is comparable to that of SimLayerKV at a \(1.53 \times \) compression rate. Our approach achieves optimal performance with minimal memory overhead, providing strong evidence for the practicality of this quantization-sparsity hybrid architecture.
In contrast, SimLayerKV requires real-time identification of layer types based on historical attention scores, making it incompatible with FlashAttention. This introduces additional computational and memory overhead, which increases latency and may cause out-of-memory issues.

\section{Effectiveness of Dynamic Retrieval}
\label{appendix:effectiveness_retrieval}
\begin{table*}[t]
\centering
\small
\setlength{\tabcolsep}{1.7pt}

\begin{tabular}{lccccccccccccccc}
\toprule
\textbf{Methods} & \textbf{Tokens} & \textbf{Qspr} & \textbf{MulFi} & \textbf{HQA} & \textbf{WMQA} & \textbf{GRpt} & \textbf{MulN} & \textbf{TREC} & \textbf{SMSM} & \textbf{TriQA} & \textbf{Repo} & \textbf{LCC } & \textbf{PsgC} & \textbf{PsgR} & \textbf{Avg.}\\
\midrule
\textit{Llama-3.1-8B} & 128k & 45.5 &53.8 & 54.7 &47.1 & 34.9 &27.5 &73.0 & 43.8 &91.6 & 56.7 & 63.4 & 7.5 & 99.5 & 53.8 \\
\midrule
\textcolor{black!60}{StreamLLM}\textsuperscript{$\ddagger$} & \textcolor{black!60}{192} & \textcolor{black!60}{21.4} & \textcolor{black!60}{31.3} & \textcolor{black!60}{46.5}  & \textcolor{black!60}{38.9} & \textcolor{black!60}{17.9} & \textcolor{black!60}{18.0} & \textcolor{black!60}{40.0}   & \textcolor{black!60}{34.4} & \textcolor{black!60}{75.7} & \textcolor{black!60}{45.7} & \textcolor{black!60}{50.7} & \textcolor{black!60}{8.0} & \textcolor{black!60}{99.0} & \textcolor{black!60}{40.6} \\

StreamLLM\textsuperscript{$\dagger$} &192 & 21.6 & 30.8 & 45.5 & 39.0 & 18.4 & 17.9 & 40.5 & 34.1 & 75.6 & 45.5 & 52.8 & 8.0 & 99.0 & 40.7 \\

\textcolor{black!60}{SnapKV}\textsuperscript{$\ddagger$} & \textcolor{black!60}{192} & \textcolor{black!60}{25.7} & \textcolor{black!60}{44.7} & \textcolor{black!60}{52.6} & \textcolor{black!60}{43.7} & \textcolor{black!60}{20.0} & \textcolor{black!60}{20.5} & \textcolor{black!60}{41.0}    & \textcolor{black!60}{39.6} & \textcolor{black!60}{89.0} & \textcolor{black!60}{48.7} & \textcolor{black!60}{57.0} & \textcolor{black!60}{8.0}  & \textcolor{black!60}{97.0} & \textcolor{black!60}{45.2} \\
SnapKV\textsuperscript{$\dagger$} & 192 & 32.4 & 47.0 & 54.6 & 44.0 & 21.9 & 22.8 & 48.0 & 40.0 & 90.3 & 51.9 & 59.9 & 8.0 & 98.0 & 47.6\\

\textcolor{black!60}{Quest}\textsuperscript{$\ddagger$} & \textcolor{black!60}{192} & \textcolor{black!60}{35.9} & \textcolor{black!60}{44.2} & \textcolor{black!60}{52.8} & \textcolor{black!60}{41.0} & \textcolor{black!60}{17.7} & \textcolor{black!60}{23.8} & \textcolor{black!60}{63.0} & \textcolor{black!60}{36.0} & \textcolor{black!60}{86.0} & \textcolor{black!60}{43.6} & \textcolor{black!60}{52.3} & \textcolor{black!60}{\textbf{8.4}} & \textcolor{black!60}{96.5} & \textcolor{black!60}{46.2} \\

Quest\textsuperscript{$\dagger$} & 192 & 39.1 & 45.1 & 52.4 & 43.4 & 21.1 & 25.6 & 65.5 & 38.7 & 88.1 & 44.8 & 52.0 & 8.1 & 97.0 & 47.8\\

\texttt{TailorKV-1} & 64(+128) & 43.4 & 53.0 & \textbf{55.3} & \textbf{46.5} & 31.3 & \textbf{27.2} & 70.0 & 42.5 & 91.6 & \textbf{54.8} & 61.8 & 7.9 & \textbf{99.0} & 52.6 \\
\texttt{TailorKV-2} & 64(+128) & \textbf{44.8} & \textbf{53.9} & 54.8 & 46.2 & \textbf{31.9} & 26.8 & \textbf{70.5} & \textbf{43.2} & \textbf{92.3} & 54.2 & \textbf{62.1} & 7.7 & \textbf{99.0} & \textbf{52.9}\\

\bottomrule
\end{tabular}
\caption{Effectiveness of dynamic retrieval. Methods marked with \textsuperscript{$\dagger$} indicate that the 0th layer of the model retains the full-precision (16-bit) KV cache, while methods marked with \textsuperscript{$\ddagger$} indicate that all layers use the same compression strategy. \texttt{TailorKV-1} and \texttt{TailorKV-2} store the KV cache as 1-bit and 2-bit, respectively, in the 0th layer.}
\label{tab:effectiveness_retrieval}
\end{table*}
\autoref{tab:effectiveness_retrieval} presents a comparison of retrieval accuracy between our sparsity-friendly layers and alternative methods, using the Llama-3.1-8B-Instruct model on the LongBench benchmark. Specifically, we retain full precision for the KV cache in the 0th layer of StreamLLM, SnapKV, and Quest, thereby preserving the global information in the 0th layer. \texttt{TailorKV-1} and \texttt{TailorKV-2} represent the quantization of the 0th layer's KV cache to 1-bit and 2-bit precision, respectively. 

The experimental results demonstrate that our retrieval method outperforms other sparse methods when the global information is preserved in the 0th layer. 
Specifically, TailorKV applies quantization to the 0th layer, whereas other methods use full precision (16-bit), and the attention calculation utilizes the same tokens from layer 1 to layer 31. 
This notable performance advantage highlights that our retrieval method effectively identifies the most important tokens, thereby minimizing the loss of crucial information.

\section{More Information on Models and Benchmarks}
\label{appendix:evaluating_tasks}

\subsection{Baselines}

In all of our experiments, we use pre-trained model weights obtained from Huggingface. These models are based on two representative attention structures: (1) MHA: including Llama-2-7B-32K-Instruct\footnote{\url{https://huggingface.co/togethercomputer/Llama-2-7B-32K-Instruct}}. (2) GQA: including Llama-3.1-8B-Instruct\footnote{\url{https://huggingface.co/meta-llama/Llama-3.1-8B-Instruct}},  Yi-6B-200K\footnote{\url{https://huggingface.co/01-ai/Yi-6B-200K}}, and Yi-9B-200K\footnote{\url{https://huggingface.co/01-ai/Yi-9B-200K}}. Detailed information about the four models can be found in \autoref{tab:model}.

To showcase the state-of-the-art performance of our method, we compare TailorKV with the following baselines:
(1) \textbf{StreamingLLM}~\cite{xiao2024efficient}: An eviction strategy that retains only the initial and most recent tokens. 
(2) \textbf{SnapKV}~\cite{li2024snapkv}: An eviction strategy that chooses clustered important KV positions. 
(3) \textbf{Quest}~\cite{tang2024quest}: A selection strategy that determines page criticality through paged key. 
(4) \textbf{PQCache}~\cite{zhang2024pqcache}: A selection strategy that retrieves Top-K tokens through vector quantization.

\subsection{Benchmarks}
\paragraph{LongBench.}
A benchmark is conducted across six categories: summarization, code completion, synthetic tasks, few-shot learning, and single/multi-document question answering. 
\autoref{tab:longbench_detail} presents detailed information on the 13 datasets in LongBench.

\paragraph{InfiniteBench.}
A benchmark designed to assess the ability of language models to process, understand, and reason with extremely long contexts (200k+ tokens).
We test the Llama3 and Yi models with context lengths of 128K and 200K, truncating inputs beyond these limits.
\autoref{tab:InfiniteBench_detail} provides details of the 9 datasets in InfiniteBench.

\begin{table*}[p]
\small
\centering
\begin{tabular}{lccccc}
\toprule
Name &   Claimed Length & Query Heads & KV Heads & Num Layers &  $\mathbb{Q}$\\ 
\midrule
Llama-3.1-8B-Instruct & 128k & 32 & 8 & 32 & \{0\}\\
Llama-2-7B-32K-Instruct & 32k & 32 &32 & 32& \{0, 1\}\\ 
Yi-6B-200K & 200k & 32 & 4 &32 & \{0, 1\}\\ 
Yi-9B-200K & 200k & 32 & 4 &48 & \{0, 1\}\\
\bottomrule
\end{tabular}
\caption{Details of models. $\mathbb{Q}$ denotes the quantization-friendly layer.}
\label{tab:model}
\end{table*}

\begin{table*}[p]
\small
\centering
\begin{tabular}{lccccc}
\toprule
Label & Task & Capability & Metric & Avg len  & \#data\\
\midrule
Qspr & Qasper & Single-Doc. QA (SD.QA) & F1 & 3,619  & 200  \\
MulFi & MultiFieldQA-en & Single-Doc. QA (SD.QA) & F1 & 4,559  & 150  \\
HQA & HotpotQA & Multi-Doc. QA (MD.QA) & F1 & 9,151  & 200  \\
WMQA & 2WikiMultihopQA & Multi-Doc. QA (MD.QA)& F1 & 4,887  & 200\\
GRpt & GovReport & Summarization (Summ)& Rouge-L & 8,734  & 200  \\
MulN & MultiNews & Summarization (Summ)& Rouge-L & 2,113  & 200 \\
TREC  & TREC & Few-shot Learning (FS.L)& Accuracy (CLS) & 5,177  & 200\\
SMSM & SAMSum & Few-shot Learning (FS.L)& Rouge-L & 6,258  & 200 \\
TriQA & TriviaQA & Few-shot Learning (FS.L)& F1 & 8,209  & 200 \\
Lcc & LCC & Code Completion (Code) & Edit Sim & 1,235   & 500\\
Repo & RepoBench-P & Code Completion (Code) & Edit Sim & 4,206  & 500  \\ 
PsgC & PassageCount & Synthetic (Synth)& Accuracy (EM) & 11,141  & 200 \\
PsgR & PassageRetrieval-en & Synthetic (Synth) & Accuracy (EM) & 9,289  & 200  \\
\bottomrule
\end{tabular}%
 \caption{Details of LongBench.}
 \label{tab:longbench_detail}
\end{table*}

\begin{table*}[p]
\small
\centering
\begin{tabular}{lcccccc}
\toprule
Label & Task & Context & Capability & Metric & Avg len  & \#Examples\\ 
\midrule
R.PK & Retrieve.PassKey & Fake Book & Retrieve (Retr)& Accuracy & 122.4k  & 590  \\
R.Num & Retrieve.Number & Synthetic & Retrieve (Retr) & Accuracy & 122.4k  & 590  \\
En.Dia  & En.Dia & Script & Dialogue (Dia) & Accuracy & 103.6k  & 200 \\
Sum  & En.Sum & Fake Book &  Novel & Rouge-L-Sum & 171.5k  & 103  \\
En.MC & En.MC & Fake Book & Novel & Accuracy & 184.4k  & 229  \\
En.QA & En.QA & Fake Book & Novel& QA F1 Score & 192.6k  & 351  \\
Zh.QA  & Zh.QA & New Book &  Novel&  QA F1 Score & 2068.6k  & 175  \\
Math.F  & Math.Find & Synthetic & Math &Accuracy & 87.9k  & 350  \\
Code.D  & Code.Debug & Code Document & Code & Accuracy & 114.7k  & 394  \\
\bottomrule
\end{tabular}%
\caption{Details of InfiniteBench.}
\label{tab:InfiniteBench_detail}
\end{table*}

\begin{table*}[p]
\small
\centering
\begin{tabular}{lcc}
\toprule
Label & Task & Category\\ 
\midrule
N-S1  & Single NIAH & Retrieval  \\
N-S2 & Single NIAH & Retrieval  \\
N-S3 & Single NIAH & Retrieval  \\
N-MK1 & Multi-keys NIAH & Retrieval  \\
N-MK2 & Multi-keys NIAH & Retrieval  \\
N-MK3 & Multi-keys NIAH & Retrieval  \\
N-MV  & Multi-values NIAH & Retrieval  \\
N-MQ & Multi-queries NIAH & Retrieval  \\
VT &  Variable Tracking &  Multi-hop Tracing  \\
CWE &  Common Words & Aggregation  \\
FWE &  Frequent Words Extraction & Aggregation  \\
QA-1 &  Question Answering &  Question Answering  \\
QA-2 &  Question Answering &  Question Answering  \\
\bottomrule
\end{tabular}%
\caption{Details of RULER.}
\label{tab:ruler_detail}
\end{table*}

\paragraph{RULER.}

A benchmark intended to assess the long-context modeling capabilities of language models, covering question answering, retrieval, aggregation, and multi-hop tracing.
This benchmark consists of 13 representative tasks, with sequence lengths ranging from 4K to 128K. For each task, we employed 25 samples. Detailed information is provided in \autoref{tab:ruler_detail}.

\section{Detailed Results}

\subsection{Accuracy on Long Context Tasks}
\label{appendix:Detailed_Results}

\autoref{tab:longbench} and \autoref{tab:infinitebench} present experimental results for LongBench and InfiniteBench. \autoref{tab:appendix_ruler} shows accuracy results for sequence lengths of 64k and 128k on RULER. 

\subsection{Efficiency Results}
\label{appendix:detail_latency}

In \autoref{tab:detail_latency}, we present the end-to-end latency for Llama-3.1-8B-Instruct, Llama-2-7B-32K-Instruct, Yi-6B-200K, and Yi-9B-200K. 
The results indicate that our method achieves efficiency closest to that of the original model.

\begin{table}[thb!]
\centering
\small
\setlength{\tabcolsep}{5pt}
\begin{tabular}{lccccc}
\toprule
\textbf{Method} & \textbf{16k} & \textbf{32k} & \textbf{64k}  & \textbf{96k} & \textbf{128k} \\
\midrule
    \multicolumn{6}{c}{Llama-3.1-8B-Instruct}
    \\
    \midrule
Full Cache & 0.024 & 0.033 & 0.050  & 0.062 & 0.082\\
OffloadCach & 0.124 & 0.227 & 0.435 & 0.743 & 0.992 \\
PQCache & 0.104 & 0.105 & 0.108 & 0.108 & 0.110\\
TailorKV & 0.045 & 0.047  & 0.054 & 0.054 & 0.056 \\
\midrule
    \multicolumn{6}{c}{Llama-2-7B-32K-Instruct}
    \\
    \midrule
Full Cache &0.045 & 0.077 & 0.140  & OOM   & OOM  \\
OffloadCach & 0.433 & 0.838 & 1.767 & 3.253 & 4.468\\
PQCache &  0.108 & 0.111 & 0.112 & 0.115 & 0.120\\
TailorKV & 0.041 & 0.062  & 0.098 & 0.132 & 0.170  \\
\midrule
    \multicolumn{6}{c}{Yi-6B-200K}
    \\
    \midrule
Full Cache & 0.019 & 0.021 & 0.029 & 0.036 & 0.044\\
OffloadCach & 0.066 & 0.118 & 0.221 & 0.325 & 0.430\\
PQCache & 0.085 & 0.087 & 0.090 & 0.092 & 0.094\\
TailorKV & 0.041 & 0.042 & 0.046 & 0.049 & 0.056 \\
\midrule
    \multicolumn{6}{c}{Yi-9B-200K}
    \\
    \midrule
Full Cache & 0.029 & 0.032 & 0.043 & 0.055 & 0.070 \\
OffloadCach & 0.102 & 0.205 & 0.417 & 0.626 & 0.843\\
PQCache & 0.130 & 0.138 & 0.139 & 0.144 & 0.150\\
TailorKV & 0.066 & 0.067 & 0.072 & 0.076 & 0.079 \\
\bottomrule
\end{tabular}
\caption{Decoding latency(s) on A100 (80G).}
\label{tab:detail_latency}
\end{table}

\begin{table*}[p]

\setlength{\tabcolsep}{1.7pt}
\small
\centering

\begin{tabular}{lcccccccccccccccc}

\toprule

\multirow{3}{*}{Method} &\multirow{3}{*}{Tokens} & \multicolumn{2}{c}{SD.QA} & \multicolumn{2}{c}{MD.QA}& \multicolumn{2}{c}{Summ}& \multicolumn{3}{c}{FS.L}& \multicolumn{2}{c}{Code} & \multicolumn{2}{c}{Synth} &\multirow{3}{*}{Avg.} \\

\cmidrule(lr){3-4}\cmidrule(lr){5-6}\cmidrule(lr){7-8}\cmidrule(lr){9-11}\cmidrule(lr){12-13}\cmidrule(lr){14-15}
&& {Qspr} & {MulFi} & {HQA} & {WMQA} & {GRpt} & {MulN} & {TREC} & {SMSM} & {TriQA} & {Repo} & {LCC} & {PsgC} & {PsgR} \\

\midrule
 \rowcolor{Blue}
  \textit{Llama-3.1-8B} & 128k & 45.5 &53.8 & 54.7 &47.1 & 34.9 &27.5 &73.0 & 43.8 &91.6 & 56.7 & 63.4 & 7.5 & 99.5 & 53.8  \\

  StreamLLM & 192 & 21.4 & 31.3 & 46.5  & 38.9 & 17.9 & 18.0 & 40.0   & 34.4 & 75.7 & 45.7 & 50.7 & 8.0 & 99.0 & 40.6\\

 SnapKV & 192& 25.7 & 44.7 & 52.6 & 43.7 & 20.0 & 20.5 & 41.0    & 39.6 & 89.0 & 48.7 & 57.0 & 8.0  & 97.0 & 45.2\\

 Quest & 192& 35.9 & 44.2 & 52.8 & 41.0 & 17.7 & 23.8 & 63.0 & 36.0 & 86.0 & 43.6 & 52.3 & \textbf{8.4} & 96.5 & 46.2 \\

  PQCache & 192& \textbf{45.6} & 51.2 & 53.8 & 45.3 & 29.0 & 25.9 & 69.5 & 41.2 & 91.3 & 54.1 & 58.5 & 8.2 & 99.0 & 51.7 \\

\rowcolor{pink!20}
 \texttt{TailorKV-1}  & 64(+128)& 43.4 & 53.0 & \textbf{55.3} & \textbf{46.5} & 31.3 & \textbf{27.2} & 70.0 & 42.5 & 91.6 & \textbf{54.8} & 61.8 & 7.9 & \textbf{99.0} & 52.6\\
 \rowcolor{pink!20}
  \texttt{TailorKV-2}  & 64(+128)& 44.8 & \textbf{53.9} & 54.8 & 46.2 & \textbf{31.9} & 26.8 & \textbf{70.5} & \textbf{43.2} & \textbf{92.3} & 54.2 & \textbf{62.1} & 7.7 & \textbf{99.0} &\textbf{52.9}\\
\midrule

 \rowcolor{Blue}
\textit{Yi-9B}&200k& 38.4 &34.9 & 52.7 &36.7 & 31.0 &26.7 &77.0 & 14.9 &90.0&67.4 & 71.9 & 2.5 & 67.5 & 47.0\\

StreamLLM& 192& 22.3 & 20.4 & 36.7 & 30.6 & 11.8  & 10.3 & 45.5  & 9.2   & 77.7 & 49.6 & 54.1 & 3.5   & 26.0 & 30.6 \\

  SnapKV & 192& 26.7 & 23.3 & 44.2 & 33.4 & 11.5 & 12.3  & 44.5  & 13.4 & 89.1 & 56.6 & 62.8 & 1.6  & 36.0 & 35.0\\
 
 Quest & 192& 32.4 & 26.0 & 44.2 & 31.5 & 14.3 & 16.5 & 73.0 & 13.4 & 86.2 & 55.6 & 63.7 & 3.9 & 47.5 & 39.1 \\
  PQCache & 192& 36.9 & 27.9 & 47.6 & 35.5 & 19.3 & 19.2 & 74.0 & 12.2 & \textbf{89.6} & 62.0 & 66.8 & \textbf{4.6} & \textbf{51.0} & 42.0   \\

 \rowcolor{pink!20}
\texttt{TailorKV-1}  & 64(+128)& \textbf{37.7} & \textbf{38.2} & \textbf{52.8} & \textbf{35.8} & \textbf{29.8} & 24.9 & \textbf{76.0} & \textbf{15.2} & 89.5 & \textbf{64.3} & 68.4 & 3.5 & 45.0 & \textbf{44.7} \\
 \rowcolor{pink!20}
\texttt{TailorKV-2}  & 64(+128)& 37.6 & 33.6 & 52.6 & 34.4 & 29.4 & \textbf{25.3} & \textbf{76.0} & 15.0 & 89.3 & 63.5 & \textbf{68.5} & 3.0 & 44.0 & 44.0 \\

 \midrule

 \rowcolor{Blue}
\textit{Yi-6B}& 200k & 25.4 &39.5 & 14.8 &15.8 & 2.8 &0.01 &72.5 & 7.7 &69.7 & 58.5 & 61.2 & 3.5 & 15.5 & 29.7\\

StreamLLM& 192& 11.0 & 29.0 & 11.1 & 12.1 & 3.1  & 0.1  & 41.5  & 5.1  & 55.5 & 43.1 & 46.2 & 3.0     & 5.0 & 20.4\\

 SnapKV & 192& 14.9 & 33.5 & 13.1 & 13.0 & \textbf{3.3}  & 0.01  & 45.0    & 6.6  & 63.8 & 48.8 & 53.7 & 3.0     & 4.5 & 23.3\\
 
 Quest & 192 & 19.9 & 33.2 & 11.8 & 13.2 & 0.5 & 0.1 & 67.0 & \textbf{9.4} & 64.5 & 50.2 & 53.7 & \textbf{3.5} & \textbf{13.5} & 26.2\\
 
PQCache & 192& 24.1 & 36.8 & 13.6 & 15.3 & 1.3 &0.01 & 68.5  & 7.6  & 68.1 & 54.1 & 57.4 & 2.5  & 5.5 & 27.3 \\

\rowcolor{pink!20}
 \texttt{TailorKV-1}  & 64(+128)& \textbf{24.5} & 40.5 & \textbf{15.6} & 15.3 & 2.7 & \textbf{0.1} & \textbf{72.0} & 8.5 & \textbf{68.3} & \textbf{55.2} & 56.5 & 2.5 & 5.5 & \textbf{28.3}\\
\rowcolor{pink!20}
 \texttt{TailorKV-2}  & 64(+128)& 24.1 & \textbf{40.8} & 15.2 & \textbf{15.5} & 3.0 & 0.01 & \textbf{72.0} & 8.6 & 66.7 & 54.8 & \textbf{57.9} & 2.5 & 5.5 & 28.2\\
\bottomrule
 
\end{tabular}

\caption{Results on \textbf{LongBench}~\cite{bai-etal-2024-longbench} of different methods. }
\label{tab:longbench}
\end{table*}

\begin{table*}[p]
    \small
    \centering
    \setlength{\tabcolsep}{2pt}

    \begin{tabular}{lccccccccccc}
    \toprule
        \textbf{Methods} & \textbf{Tokens} & \textbf{R.PK} & \textbf{R.Num}  & \textbf{En.Dia} & \textbf{Sum}  & \textbf{En.MC} & \textbf{En.QA} & \textbf{Zh.QA} &  \textbf{Math.F} & \textbf{Code.D} & \textbf{Avg.} \\

    \midrule
     \rowcolor{Blue}
    \textit{Llama-3.1-8B} & 128K & 100.0 & 99.3  & 19.0 & 26.8 & 65.9 & 14.8 & 13.3 & 34.0 & 22.8 & 44.0 \\
    
    StreamLLM & 1024 & 3.3 & 3.0  & 7.0 & 12.7 & 66.3 & 5.9 & 9.7 & 34.0 & 22.8 & 18.3 \\

    SnapKV & 1024 & 100.0 & \textbf{93.2}  & 9.5 & 22.4 & 65.5 & 10.4 & 11.3 & 34.0 & 22.8 & 41.0 \\
    
    Quest & 1024 & 100.0 & 28.9  & 14.0 & 12.2 & \textbf{69.8} & 9.2 & 11.4 & 34.0 & \textbf{25.1} &  33.8\\
    
    PQCache  & 1024 & 8.6 & 2.5  & 15.0 & 18.9 & 65.9 & 12.6 & 12.6 & 34.0 & 23.3 & 21.5 \\ 
\rowcolor{pink!20}
 \texttt{TailorKV-1}  & 128+(896) & 99.8 & 73.2  & 18.0 & 22.8 & 66.4 & 13.6 & \textbf{13.0} & \textbf{34.0} & 22.8 & 40.4 \\
\rowcolor{pink!20}
 \texttt{TailorKV-2}  & 128+(896) & \textbf{100.0} & 89.4  & \textbf{18.5} & \textbf{24.1} & 66.8 & \textbf{14.4} & 12.9 & \textbf{34.0} & 22.8 & \textbf{42.6} \\
    \midrule
    \rowcolor{Blue}
    \textit{Yi-9B} & 200K & 100.0 & 100.0  & 2.5 & 8.2 & 65.0 & 10.8 & 16.7 & 23.4 & 26.3 & 39.2  \\

    StreamLLM & 1024 & 2.5 & 0.5  & 2.5 & 6.4 & 66.8 & 8.8 & 15.0 & 23.7 & 21.3 & 16.4\\

    SnapKV & 1024 & 99.8 & 18.3  & 3.0 & 8.6 & \textbf{67.6} & 8.4 & 14.9 & 22.5 & \textbf{26.6} & 30.0 \\
    
    Quest & 1024 &100.0 & 96.9  & 4.0 & 3.4 & 58.5 & 12.5 & 12.7 & 18.2 & 18.7 &  36.1\\

    PQCache & 1024 & 9.6 & 5.9  & 2.0 & \textbf{8.7} & 66.3 & 11.2 & 14.9 & 22.2 & 25.6 & 18.5 \\
    
\rowcolor{pink!20}
 \texttt{TailorKV-1}  & 128+(896) & \textbf{100.0} & 98.3  & 2.5 & 7.3 & 64.2 & \textbf{15.1} & \textbf{19.7} & \textbf{24.0} & 21.3 & 39.2  \\
\rowcolor{pink!20}
 \texttt{TailorKV-2}  & 128+(896) & \textbf{100.0} & \textbf{99.7}  & \textbf{4.5} & 8.2 & 65.1 & 11.2 & 16.6 & \textbf{24.0} & 24.9 & \textbf{39.4}  \\

    \midrule
    \rowcolor{Blue}
    \textit{Yi-6B} & 200K & 100.0 & 98.4  & 1.0 & 3.4 & 53.3 & 18.2 & 26.0 & 6.8 & 26.9 & 37.1 \\
    
    StreamLLM & 1024 & 3.3 & 0.6  & 0.0 & 3.9 & 52.8 & 10.2 & 20.2 & 4.8 & 25.8 & 13.5 \\

    SnapKV & 1024 & 100.0 & 11.5  & 2.5 & 3.1 & 53.2 & 14.1 & 22.2 & 4.5 & \textbf{26.9}  & 26.4\\
    
    Quest & 1024 & 100.0 & \textbf{99.1}  & 3.0 &  3.6 & 52.4 & 13.4 & 18.8 & 5.1 & 26.6  & 35.8 \\

    PQCache & 1024 & 10.1 & 3.7  & 1.5 & 4.4 & 51.5 & 16.4 & \textbf{26.2} & 5.7 & \textbf{26.9} & 16.3 \\
    
\rowcolor{pink!20}
 \texttt{TailorKV-1}  & 128+(896) & \textbf{100.0} & 97.5  & 2.5 & \textbf{4.4} & \textbf{53.3} & 17.4 & 26.0 & 7.7 & 26.4 & 37.2 \\
 \rowcolor{pink!20}
 \texttt{TailorKV-2}  & 128+(896) & \textbf{100.0} & 97.0  & \textbf{3.0} & 4.0 & \textbf{53.3} & \textbf{18.0} & 25.8 & \textbf{8.0} & 26.7 & \textbf{37.3} \\
    \bottomrule
    \end{tabular}
    
    \caption{Results on \textbf{InfiniteBench}~\cite{zhang-etal-2024-bench} of different methods.}
    \label{tab:infinitebench}
\end{table*}

\begin{table*}[p]
    \small
    \centering
    \setlength{\tabcolsep}{3pt}

    \begin{tabular}{lcccccccccccccc}
    \toprule
        \textbf{Methods} &  \textbf{N-S1} & \textbf{N-S2} & \textbf{N-S3}  & \textbf{N-MK1} & \textbf{N-MK2} & \textbf{N-MK3} & \textbf{N-MV} & \textbf{N-MQ} &  \textbf{VT} & \textbf{CWE} & \textbf{FWE} & \textbf{QA-1} & \textbf{QA-2} & \textbf{Avg.} \\

    \midrule
      \multicolumn{14}{c}{\textbf{Sequence Length = 64k}} \\
    \midrule
     \rowcolor{Blue}
    \textit{Llama-3.1-8B} & 100.0 & 100.0 & 100.0 & 100.0 & 100.0 & 96.0 & 99.0 & 100.0 & 100.0 & 14.8 & 92.0 & 60.0 & 52.0 & 85.6 \\
    
    StreamLLM & 8.0 & 4.0 & 0.0 & 8.0 & 0.0 & 0.0 & 5.0 & 3.0 & 2.4 & 0.8 & 72.0 & 28.0 & 28.0 & 12.2 \\

    SnapKV & 96.0 & 84.0 & 0.0 & 88.0 & 32.0 & 0.0 & 40.0 & 69.0 & 74.4 & 0.8 & 41.3 & 56.0 & 48.0 & 48.4\\
    
    Quest & 88.0 & 100.0 & 60.0 & 92.0 & 72.0 & 0.0 & 93.0 & 90.0 & 80.0 & 8.4 & 70.6 & 52.0 & 52.0 & 66.0\\
    
    PQCache  &36.0 & 60.0 & 12.0 & 68.0 & 48.0 & 4.0 & 16.0 & 37.0 & 52.8 & 0.0 & \textbf{73.3} & 56.0 & 48.0 & 39.2 \\
\rowcolor{pink!20}
 \texttt{TailorKV-1}  & \textbf{100.0} & \textbf{100.0} & 96.0 & \textbf{100.0} & \textbf{96.0} & 28.0 & \textbf{99.0} & \textbf{100.0} & 85.6 & 18.4 & 57.3 & 56.0 & 48.0 & 75.7 \\
 \rowcolor{pink!20}
  \texttt{TailorKV-2}  & \textbf{100.0} & \textbf{100.0}& \textbf{100.0}& \textbf{100.0}& \textbf{96.0}& \textbf{68.0}& 97.0& 98.0& \textbf{88.0}& \textbf{19.6}& 62.7& \textbf{60.0}& \textbf{56.0} & \textbf{80.4}\\
    \midrule
    \rowcolor{Blue}
    \textit{Yi-9B} & 100.0 & 100.0 & 100.0 & 100.0 & 92.0 & 48.0 & 61.0 & 88.0 & 12.8 & 15.6 & 88.0 & 32.0 & 48.0 & 68.1 \\

    StreamLLM & 0.0 & 4.0 & 0.0 & 4.0 & 0.0 & 0.0 & 1.0 & 0.0 & 0.0 & 1.2 & \textbf{74.6} & 16.0 & 28.0 & 9.9  \\

    SnapKV & 80.0 & 28.0 & 0.0 & 20.0 & 4.0 & 0.0 & 11.0 & 11.0 & 22.4 & 5.6 & 48.0 & 24.0 & 44.0 & 22.9 \\
    
    Quest & 68.0 & 92.0 & 20.0 & 68.0 & 40.0& 0.0& 24.0& 42.0 & 16.0& 10.8 & 62.6& 28.0 & 36.0 & 39.0\\

    PQCache & 32.0 & 56.0 & 4.0 & 36.0 & 16.0 & 0.0 & 7.0 & 39.0 & 31.2 & 6.0 & 66.6 & 20.0 & 36.0 & 26.9\\
    
\rowcolor{pink!20}
 \texttt{TailorKV-1}  & \textbf{100.0} & \textbf{100.0} & \textbf{92.0} & \textbf{100.0} & \textbf{84.0} & \textbf{28.0} & 53.0 & 89.0 & 6.4  & 32.0 & 49.3 & \textbf{32.0} & \textbf{48.0} & 62.6 \\
\rowcolor{pink!20}
 \texttt{TailorKV-2}  & \textbf{100.0} & \textbf{100.0} & \textbf{92.0}  & \textbf{100.0} & \textbf{84.0} & \textbf{28.0} & \textbf{62.0} & \textbf{90.0} & \textbf{42.4} & \textbf{33.6} & 48.0 & 28.0 & \textbf{48.0} & \textbf{65.8} \\
    \midrule
    \rowcolor{Blue}
    \textit{Yi-6B} & 100.0 & 100.0 & 100.0 & 96.0 & 56.0 & 24.0 & 39.0 & 76.0 & 24.8 & 0.8 & 73.3 & 32.0 & 24.0 & 57.3 \\
    
    StreamLLM & 0.0 & 0.0 & 0.0 & 8.0 & 0.0 & 0.0 & 3.0 & 0.0 & 0.0 & 0.4 & 62.6 & 20.0 & 16.0 & 8.5 \\

    SnapKV & 88.0 & 4.0 & 0.0 & 16.0 & 0.0 & 0.0 & 5.0 & 7.0 & 15.2 & 0.0 & \textbf{65.3} & 28.0 & 20.0 & 19.1\\
    
    Quest & 72.0 & 84.0 & 0.0 & 52.0 & 20.0 & 0.0 & 28.0 & 30.0 & 20.0 & \textbf{1.6} & 56.0 & 24.0 & 20.0 & 31.3\\

    PQCache & 16.0 & 20.0 & 0.0 & 28.0 & 8.0 & 0.0 & 5.0 & 3.0 & 10.4 & 0.0 & 50.6 & 24.0 & 24.0 & 14.5\\
    
\rowcolor{pink!20}
 \texttt{TailorKV-1} & \textbf{100.0} & \textbf{100.0} & \textbf{100.0} & 96.0 & 12.0 & 24.0 & \textbf{41.0} & 65.0 & 28.8 & 1.2 & 58.7 & 28.0 & \textbf{24.0} & 52.2 \\
 \rowcolor{pink!20}
 \texttt{TailorKV-2} & \textbf{100.0} & \textbf{100.0} & \textbf{100.0} & \textbf{100.0} & \textbf{24.0} & \textbf{28.0} & 40.0 & \textbf{67.0} & \textbf{42.4} & 0.8 & 57.3 & \textbf{32.0} & \textbf{24.0} & \textbf{55.1} \\

    \midrule
      \multicolumn{14}{c}{\textbf{Sequence Length = 128k}} \\
    \midrule
     \rowcolor{Blue}
    \textit{Llama-3.1-8B} & 100.0 & 100.0 & 100.0 & 100.0 & 88.0 & 64.0 & 96.0 & 98.0 & 95.2 & 1.6  & 66.6 & 64.0 & 36.0 & 77.6\\
    
    StreamLLM & 0.0  & 4.0  & 0.0  & 0.0  & 4.0  & 0.0  & 5.0  & 4.0  & 0.0  & 0.4  & 9.3 & 24.0 & 20.0 & 5.4\\

    SnapKV & 100.0 & 84.0 & 0.0  & 84.0 & 24.0 & 0.0  & 19.0 & 38.0 & 65.6 & 0.0  & 28.0  & 48.0 & 32.0 & 40.2\\
    
    Quest & 80.0 & 68.0 & 0.0 & 88.0 & 48.0 & 0.0 & 66.0 & 71.0 & 59.2 & 0.4 & 52.0 & 48.0 & 28.0  & 46.8\\
    
    PQCache  &0.0 & 8.0 & 0.0 & 4.0 & 8.0 & 0.0 & 2.0 & 3.0 & 0.8 & 0.0 & \textbf{66.6} & 40.0 & 32.0 & 12.6\\
\rowcolor{pink!20}
 \texttt{TailorKV-1}  & 92.0 & \textbf{92.0} & \textbf{100.0} & \textbf{100.0} & \textbf{64.0} & 0.0 & 93.0 & \textbf{98.0} & 67.2 & \textbf{0.4} & 16.0 & 60.0 & \textbf{40.0} & 63.3 \\
 \rowcolor{pink!20}
  \texttt{TailorKV-2}  & \textbf{100.0} & \textbf{92.0} & \textbf{100.0} & \textbf{100.0}& \textbf{64.0}& \textbf{16.0}& \textbf{96.0}& 97.0& \textbf{85.6}& \textbf{0.4}& 40.0& \textbf{64.0}&36.0 & \textbf{68.5}\\
    \midrule
    \rowcolor{Blue}
    \textit{Yi-9B} & 100.0 & 100.0 & 100.0 & 96.0  & 80.0 & 28.0 & 69.0 & 84.0 & 10.4 & 3.6  & 89.3 & 36.0 & 36.0 & 64.0\\

    StreamLLM & 0.0  & 4.0  & 4.0  & 0.0  & 0.0  & 0.0  & 2.0  & 1.14 & 0.0  & 0.0  & \textbf{86.6} & 16.0 & 24.0 & 10.6 \\

    SnapKV & 92.0  & 12.0 & 0.0  & 20.0 & 4.0  & 0.0  & 12.0 & 4.0  & 7.2  & 2.0  & 53.3 & 20.0 & 32.0 & 19.9\\
    
    Quest & 100.0 & 84.0 & 4.0 & 72.0 & 24.0 & 0.0 & 28.0 & 28.0 & 16.8 & 0.8 & 69.3 & 24.0 & 32.0 & 37.1\\

    PQCache & 8.0 & 16.0 & 0.0 & 24.0 & 4.0 & 0.0 & 2.0 & 5.0 & 4.0 & 0.4 & 77.3 & 16.0 & 28.0 & 14.2\\
    
\rowcolor{pink!20}
 \texttt{TailorKV-1}  & \textbf{100.0} & \textbf{100.0} & \textbf{96.0} & \textbf{96.0} & 72.0 & \textbf{20.0} & 44.0 & 79.6 & 19.2 & 23.2 & 44.0 & \textbf{40.0} & \textbf{32.0} & 58.9 \\
\rowcolor{pink!20}
 \texttt{TailorKV-2}  & \textbf{100.0} & \textbf{100.0} & \textbf{96.0}  & \textbf{96.0} & \textbf{76.0} & \textbf{20.0} & \textbf{55.0} & \textbf{80.0} & \textbf{48.8} & \textbf{24.0} & 41.3 & 36.0 & \textbf{32.0} & \textbf{61.9}\\
    \midrule
    \rowcolor{Blue}
    \textit{Yi-6B} & 100.0 & 100.0 & 100.0 & 84.0  & 72.0 & 4.0  & 30.0 & 67.0 & 4.8  & 1.2  & 100.0 & 32.0 & 24.0 & 55.3\\
    
    StreamLLM & 0.0  & 4.0  & 4.0  & 0.0  & 0.0  & 0.0  & 2.0  & 1.0  & 0.0  & 0.8  & 68.0  & 20.0 & 16.0 & 8.9\\

    SnapKV & 76.0  & 0.0  & 0.0  & 16.0 & 0.0  & 0.0  & 1.0  & 4.0  & 8.0  & 0.8  & \textbf{69.3} & 20.0 & 16.0 & 16.2 \\
    
    Quest & 96.0 & 72.0 & 0.0 & 72.0 & 20.0 & 0.0 & 17.0 & 23.0 & 6.4 & 0.8 & 49.3 & 16.0 & 8.0 & 29.2 \\

    PQCache & 0.0 & 8.0 & 0.0 & 12.0 & 4.0 & 0.0 & 1.0 & 2.0 & 0.0 & 0.0 & 54.6 & 24.0 & \textbf{28.0} & 10.2 \\
    
\rowcolor{pink!20}
 \texttt{TailorKV-1} & \textbf{100.0} & \textbf{100.0} & \textbf{100.0} & \textbf{84.0} & 16.0 & 0.0 & \textbf{25.0} & 47.0 & 3.2 & 0.8 & 61.3 & \textbf{32.0} & 22.7 & 45.5 \\
 \rowcolor{pink!20}
 \texttt{TailorKV-2} & \textbf{100.0} & \textbf{100.0} & \textbf{100.0} & \textbf{84.0} & \textbf{28.0} & \textbf{4.0} & 21.0 & \textbf{48.0} & \textbf{14.4} & \textbf{1.2} & 60.0 & \textbf{32.0} & 20.0 & \textbf{47.1} \\
    
    \bottomrule
    \end{tabular}
    
    \caption{Accuracy (\%) of different methods and models on \textbf{RULER}~\cite{hsieh2024ruler} evaluated at length of 64k and 128k. The sparsity-friendly layer in TailorKV uses 128+(896) tokens, while other methods use 1024 tokens.}
    \label{tab:appendix_ruler}
\end{table*}

\section{Attention Visualization Across Models}
\label{appendix:detail_attention}
\begin{figure*}[t]
\centering
\subfloat[Visualization of attention weights on Llama-2-7B-32K-Instruct.]{
    \label{fig:llama2_maps} 
    \includegraphics[width=2\columnwidth]{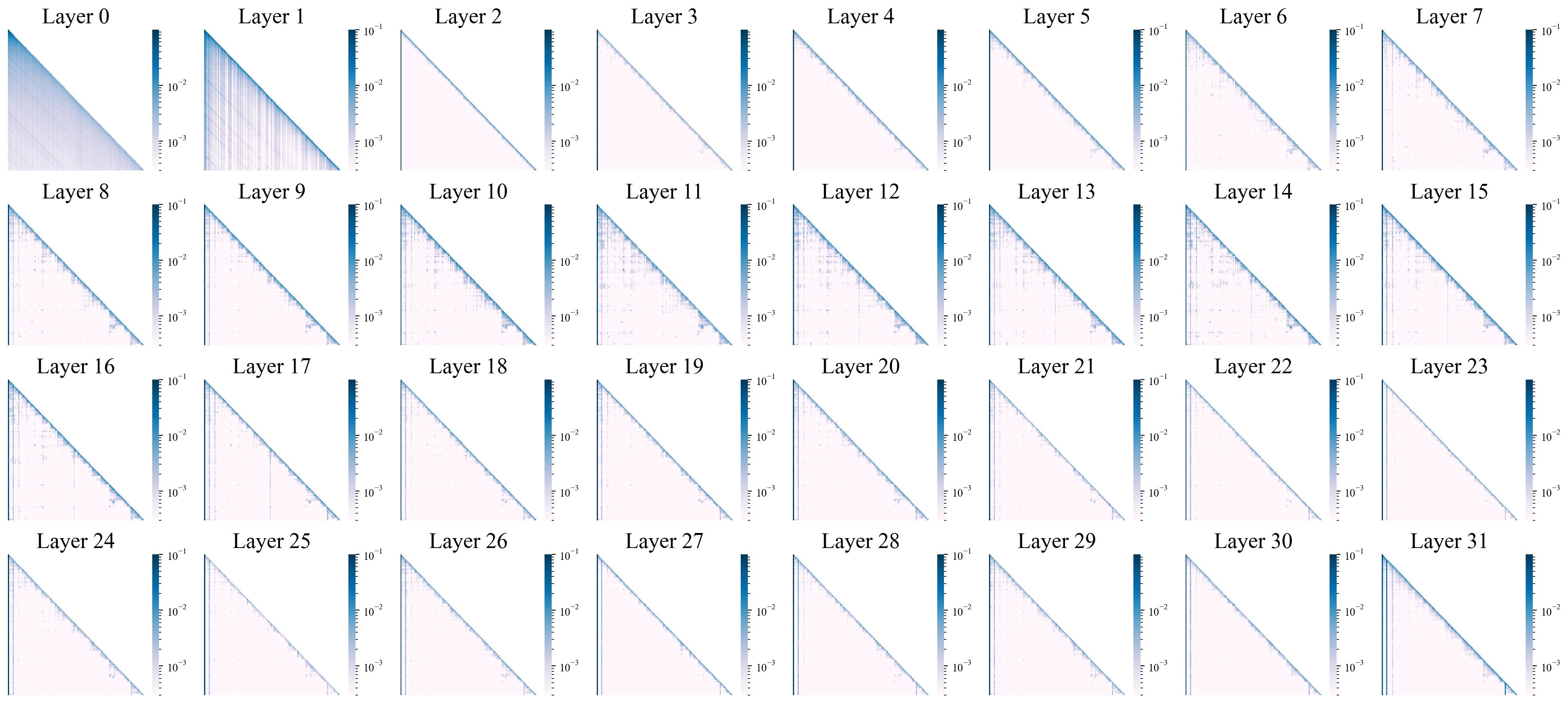}
}
\\ 
\subfloat[Visualization of attention weights on Llama-3.1-8B-Instruct.]{
    \label{fig:llama3_maps} 
    \includegraphics[width=2\columnwidth]{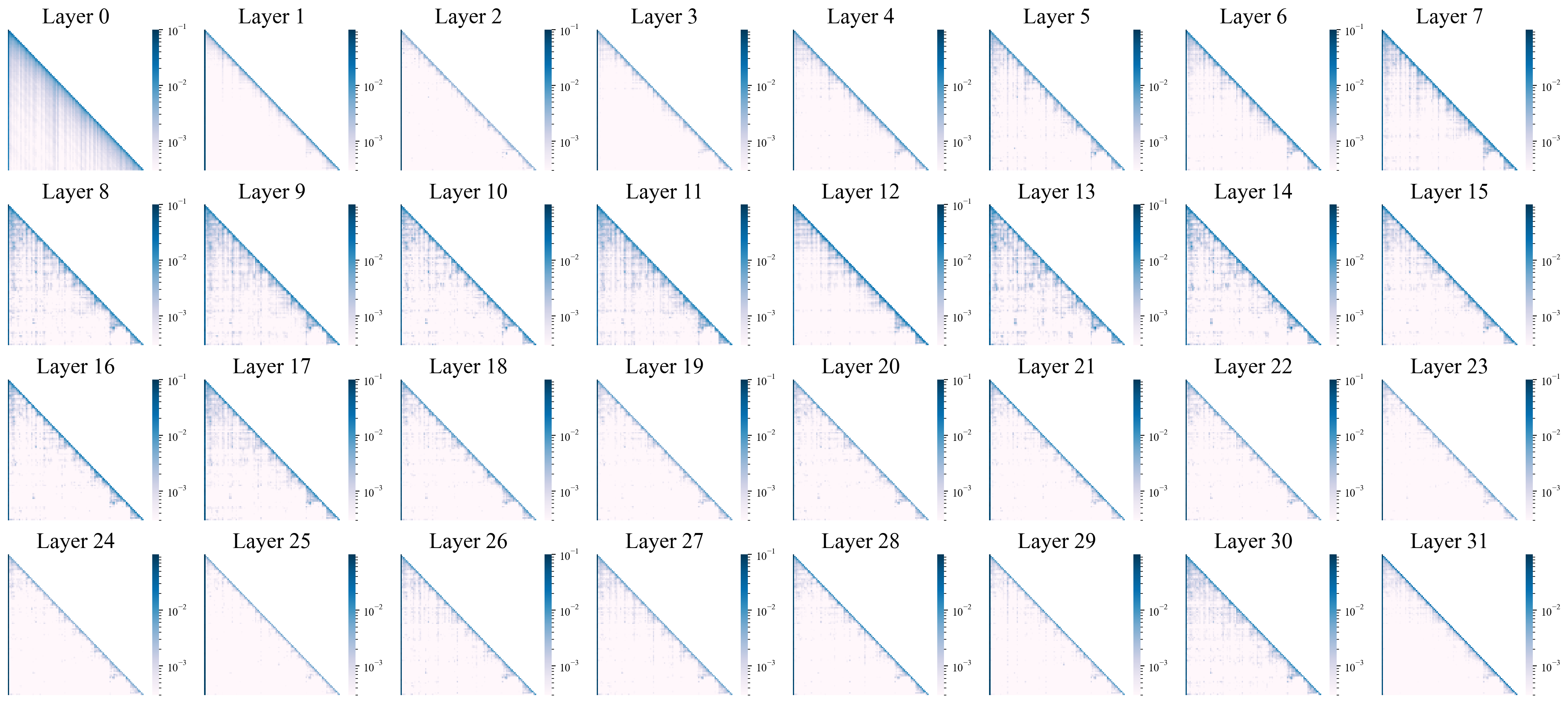}
}
\\ 
\subfloat[Visualization of attention weights on Yi-6B-200K.]{
    \label{fig:yi6_maps} 
    \includegraphics[width=2\columnwidth]{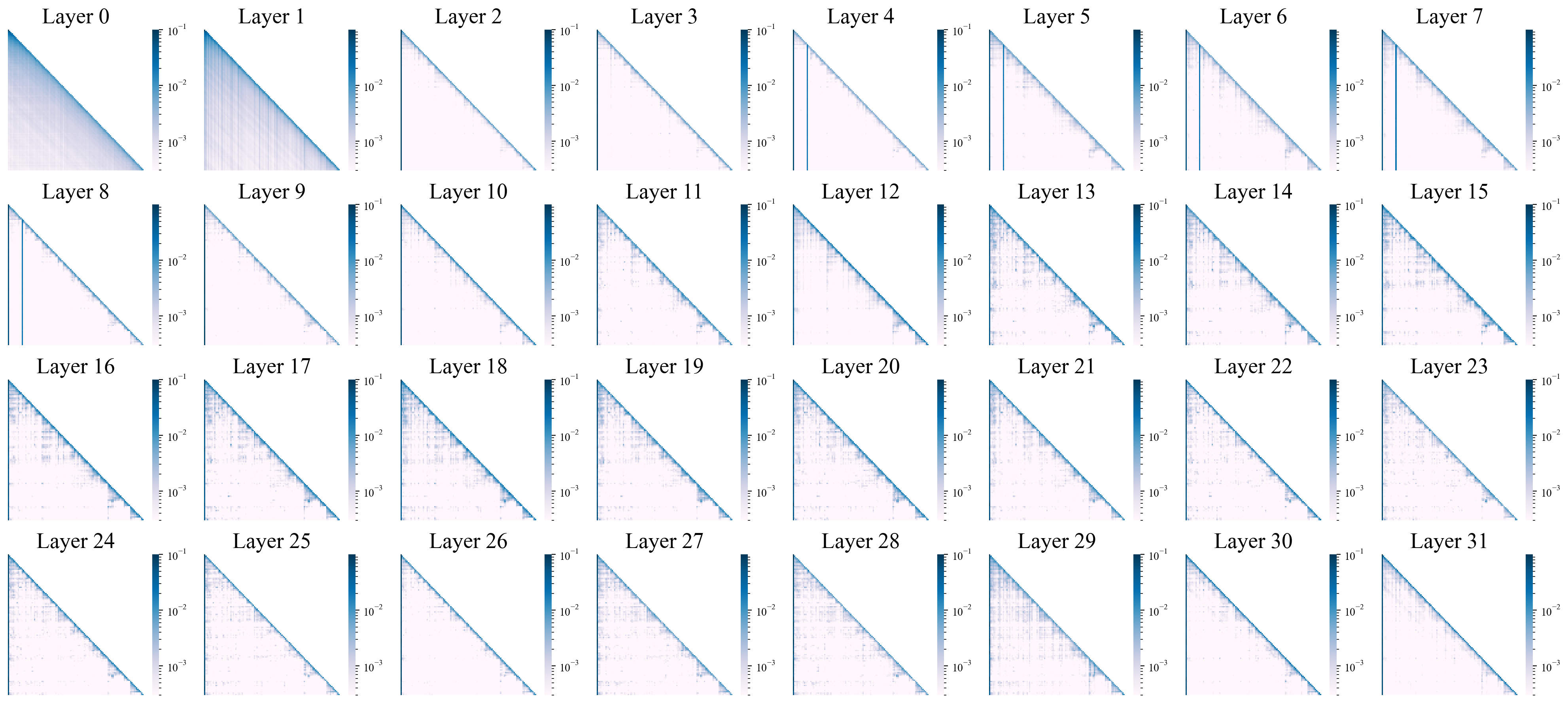}
}
\caption{Visualization of attention weights across the 2WikiMQA dataset.} 
\label{fig:detail_attention} 
\end{figure*}

As shown in \autoref{fig:detail_attention}, the attention patterns of different models closely match the results predicted by our usage metric \( \mathcal{P} \).
Specifically, the quantization-friendly layers of Llama-2-7B-32K-Instruct and Yi-6B-200K are identified as the 0th and 1st layers. In these layers, attention patterns are dense, while the other layers are sparse. 
Similarly, the quantization-friendly layer of Llama-3.1-8B-Instruct is the 0th layer, where attention pattern is dense, with sparse features in the remaining layers.

\section{Observations on QKV}
\label{appendix:detail_qkv}

\begin{figure*}[t]
  \centering
  \includegraphics[width=0.8\linewidth]{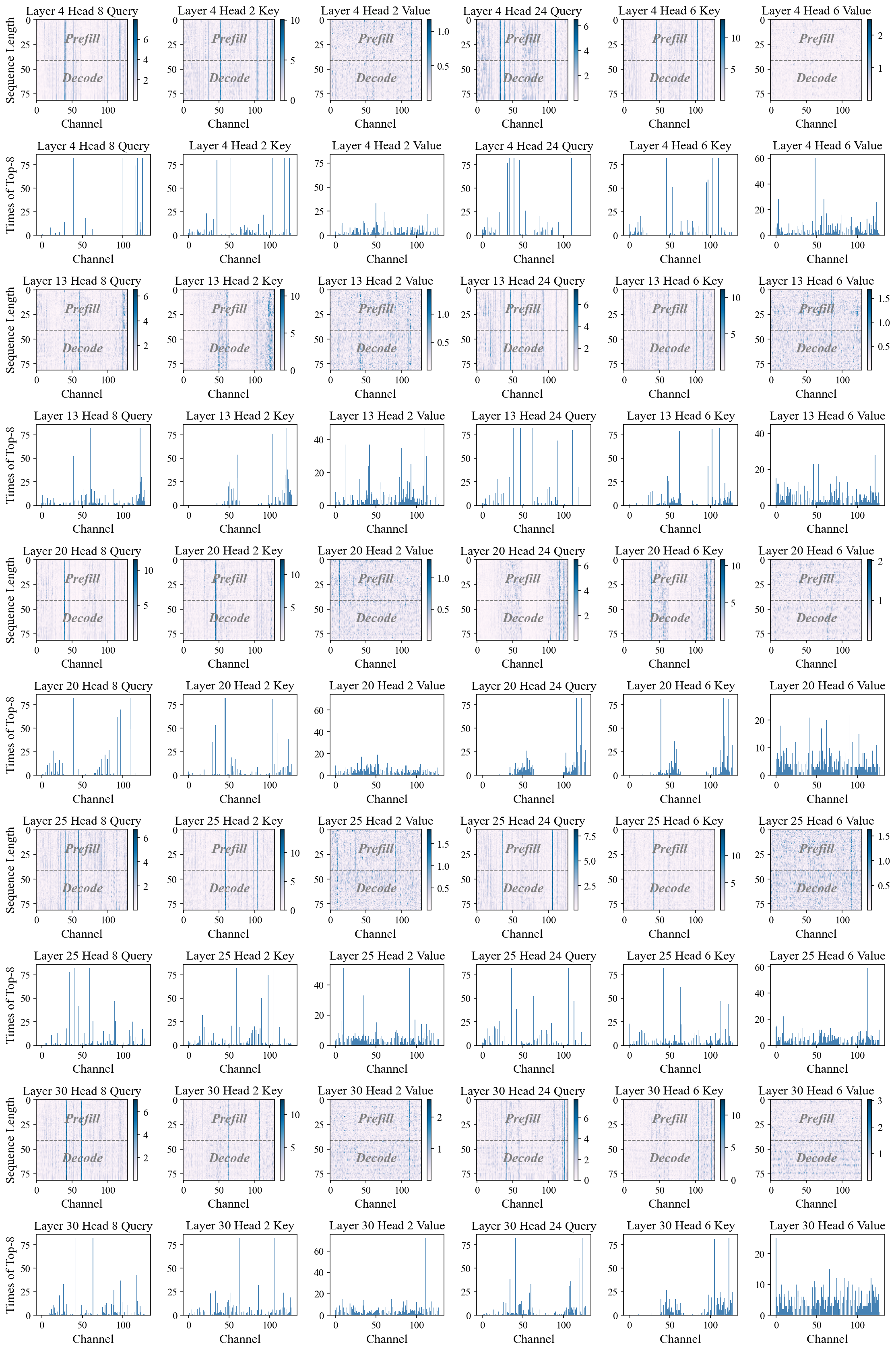} 
  \caption{Magnitude of query, key and value for Llama-3.1-8B-Instruct.}
  \label{fig:detail_qkv}
\end{figure*}

\autoref{fig:detail_qkv} illustrates the distribution patterns of queries, keys, and values across different attention heads in Llama-3.1-8B-Instruct. Although outliers appear in both the keys and queries, the locations of the outlier channels are not consistently fixed.

\end{document}